%% file: main.tex
\documentclass[runningheads]{llncs}
\usepackage{graphicx}
\usepackage{comment}
\usepackage{amsmath,amssymb} %
\usepackage{color}
\usepackage{xcolor}
\usepackage{booktabs}
\usepackage{subcaption}
\usepackage[colorlinks=true]{hyperref}
\usepackage{bm}
\usepackage{xspace}
\usepackage{dsfont}
\usepackage[frozencache]{minted}
\usepackage{graphics}

\usepackage{textcomp}

\usepackage[width=122mm,left=12mm,paperwidth=146mm,height=193mm,top=12mm,paperheight=217mm]{geometry}

\newcommand{\chk}{{\centering\checkmark}}
\newcommand{\detr}{DETR\xspace}

\newcommand{\gab}[1]{{\color{purple} gab: #1}}

\newcommand{\noobject}{\varnothing}
\newcommand{\oldnew}[2]{#2}

\makeatletter
\renewcommand\subsubsection{\@startsection{subsubsection}{4}{\z@}%
  {.5em \@plus1ex \@minus.1ex}%
  {-.5em}%
  {\normalfont\normalsize\bfseries}}
\makeatother

\include{math_commands}

\newcommand{\AP}[1]{AP\textsubscript{#1}}

\makeatletter
\DeclareRobustCommand\onedot{\futurelet\@let@token\@onedot}
\def\@onedot{\ifx\@let@token.\else.\null\fi\xspace}

\def\eg{\emph{e.g}\onedot} 
\def\ie{\emph{i.e}\onedot}

\def\wrt{w.r.t\onedot} 
\def\etal{\emph{et al}\onedot}
\makeatother

\makeatletter
\newcommand{\printfnsymbol}[1]{%
  \textsuperscript{\@fnsymbol{#1}}%
}
\makeatother

\begin{document}
\pagestyle{headings}
\mainmatter

\title{End-to-End Object Detection with Transformers}

\titlerunning{End-to-End Object Detection with Transformers}
\authorrunning{Carion et al.}
\author{
Nicolas Carion\thanks{Equal contribution} \and Francisco Massa\printfnsymbol{1}  \and Gabriel Synnaeve \and Nicolas Usunier \and Alexander Kirillov \and Sergey Zagoruyko}
\institute{Facebook AI}

\maketitle

\begin{abstract}
	We present a new method that views object detection as a direct set prediction problem. Our approach streamlines the detection pipeline, effectively removing the need for many hand-designed components like a non-maximum suppression procedure or anchor generation that explicitly encode our prior knowledge about the task. The main ingredients of the new framework, called DEtection TRansformer or \detr, are a set-based global loss that forces unique predictions
	 via bipartite matching, %
	 and a transformer encoder-decoder architecture. Given a fixed small set of learned object queries, \detr reasons about the relations of the objects and the global image context to directly output the final set of predictions in parallel. 
	The new model is conceptually simple and does not require a specialized library, unlike many other modern detectors.
	\detr %
	demonstrates accuracy and run-time performance on par with the
  well-established and highly-optimized Faster R-CNN baseline on the challenging
  COCO object detection dataset. 
  Moreover, \detr can be easily generalized to produce panoptic segmentation in a unified manner. We show that it significantly outperforms competitive baselines.
  Training code and pretrained models are available at \url{https://github.com/facebookresearch/detr}.

\end{abstract}

\section{Introduction}

The goal of object detection is to predict a set of  bounding boxes and category labels for each object of interest.
Modern detectors address this set prediction task in an indirect way, by defining surrogate regression and classification problems on a large set of proposals~\cite{Ren2015FasterRT,Cai2019CascadeRH}, anchors~\cite{Lin2017FocalLF}, or window centers~\cite{Zhou2019objects,tian2019fcos}.
Their performances are significantly influenced by postprocessing steps to collapse near-duplicate predictions, by the design of the anchor sets and by the heuristics that assign target boxes to anchors~\cite{Zhang2019bridging}.
To simplify these pipelines, we propose a direct set prediction approach to bypass the surrogate tasks. 
This end-to-end philosophy has led to significant advances in complex structured prediction tasks such as machine translation or speech recognition, but not yet in object detection: previous attempts \cite{Stewart2015EndtoEndPD,Hosang2017LearningNS,Bodla2017SoftNMSI,rezatofighi2018deep} either add other forms of prior knowledge, or have not proven to be competitive with strong baselines on challenging benchmarks. This paper aims to bridge this gap.

We streamline the training pipeline by viewing object detection as a direct set prediction problem. We adopt an encoder-decoder architecture based on transformers \cite{Vaswani2017AttentionIA}, a popular architecture for sequence prediction. The self-attention mechanisms of transformers, which explicitly model all pairwise interactions between elements in a sequence, make these architectures particularly suitable for specific constraints of set prediction such as removing duplicate predictions. 

\begin{figure}[t]
    \centering\small
    \includegraphics[width=1.0\columnwidth]{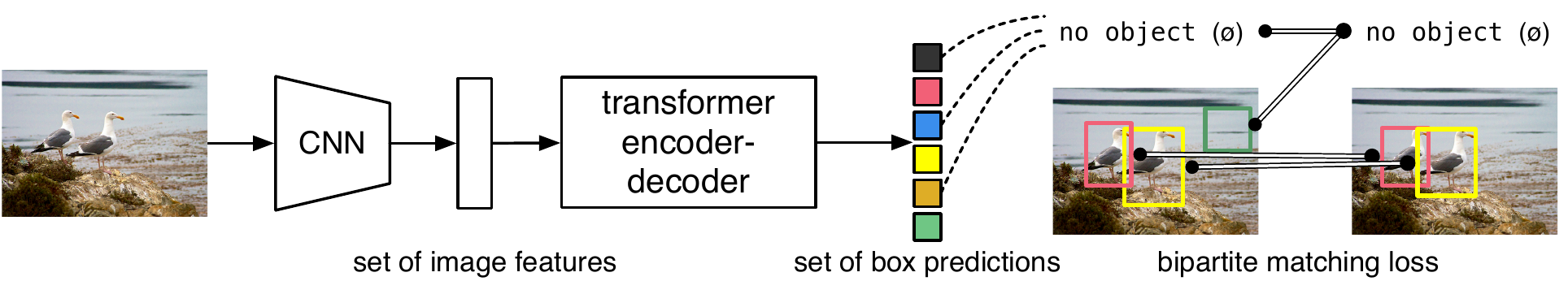}
    \caption{\detr directly predicts (in parallel)
    the final set of detections by combining a common CNN with a transformer architecture. During training, bipartite matching uniquely assigns predictions with ground truth boxes. Prediction with no match should yield a ``{\tt no object}'' ($\noobject$) class prediction.}
    \label{fig:teaser}
\end{figure}

Our DEtection TRansformer (\detr, see Figure~\ref{fig:teaser}) predicts all objects at once, and is trained end-to-end with a set loss function which performs bipartite matching between predicted and ground-truth objects.
\detr simplifies the detection pipeline by dropping multiple hand-designed components that encode prior knowledge, like spatial anchors or non-maximal suppression. 
Unlike most existing detection methods, \detr doesn't require any customized layers, and thus can be reproduced easily in any framework that contains standard CNN and transformer classes.\footnote{In our work we use standard implementations of Transformers~\cite{Vaswani2017AttentionIA} and ResNet~\cite{He2015DeepRL} backbones from standard deep learning libraries.}.

Compared to most previous work on direct set prediction, the main features of \detr are the conjunction of the bipartite matching loss and transformers with (non-autoregressive) parallel decoding \cite{oord2017parallel,Gu2017NonAutoregressiveNM,ghazvininejad2019constant,Devlin2019BERTPO}. In contrast, previous work focused on autoregressive decoding with RNNs \cite{Stewart2015EndtoEndPD,RomeraParedes2015RecurrentIS,Park2015LearningTD,ren2017end,Salvador2017RecurrentNN}. Our matching loss function uniquely assigns a prediction to a ground truth object, and is invariant to a permutation of predicted objects, so we can emit them in parallel. 

We evaluate \detr on one of the most popular object detection datasets, COCO~\cite{Lin2014coco}, against a very competitive Faster R-CNN baseline~\cite{Ren2015FasterRT}. Faster R-CNN has undergone many design iterations and its performance was greatly improved since the original publication. Our experiments show that our new model
achieves comparable performances. More precisely, \detr demonstrates significantly better performance on large objects, a result likely enabled by the non-local computations of the transformer. It obtains, however, lower performances on small objects. We expect that future work will improve this aspect in the same way the development of FPN~\cite{lin2017feature} did for Faster R-CNN.

Training settings for \detr differ from standard object detectors in multiple ways. The new model requires extra-long training schedule and benefits from auxiliary decoding losses in the transformer.
We thoroughly explore what components are crucial for the demonstrated performance.

The design ethos of \detr easily extend to more complex tasks. In our
experiments, we show that a simple segmentation head trained on top of a
pre-trained \detr outperfoms competitive baselines on Panoptic Segmentation~\cite{Kirillov2019panoptic}, a challenging pixel-level recognition task that has recently gained popularity.

\input{related}

\input{method}

\input{experiments}

\section{Conclusion}

We presented \detr, a new design for object detection systems based on transformers and bipartite matching loss for direct set prediction. The approach achieves comparable results to an optimized Faster R-CNN baseline on the challenging COCO dataset. \detr is straightforward to implement and has a flexible architecture that is easily extensible to panoptic segmentation, with competitive results. In addition, it achieves significantly better performance on large objects than Faster R-CNN, likely thanks to the processing of global information performed by the self-attention.

This new design for detectors also comes with new challenges, in particular regarding training, optimization and performances on small objects. Current detectors required several years of improvements to cope with similar issues, and we expect future work to successfully address them for \detr.

\section{Acknowledgements}

We thank Sainbayar Sukhbaatar, Piotr Bojanowski, Natalia Neverova, David Lopez-Paz, Guillaume Lample, Danielle Rothermel, Kaiming He, Ross Girshick, Xinlei Chen and the whole Facebook AI Research Paris team for discussions and advices without which this work would not be possible.

\bibliography{detection_transformer}
\bibliographystyle{splncs04}

\appendix
\newpage
\input{supplementary}

\end{document}

%% file: math_commands.tex
\renewcommand{\Re}{\mathbb{R}}
\newcommand{\hy}{\hat{y}}
\newcommand{\hb}{\hat{b}}
\newcommand{\hp}{\hat{p}}

\renewcommand{\Sigma}{\mathfrak{S}}

\newcommand{\mhattn}{\text{\rm mh-attn}}
\newcommand{\mhsattn}{\text{\rm mh-s-attn}}
\newcommand{\attn}{\text{\rm attn}}
\newcommand{\xq}{X_{\rm q}} %
\newcommand{\xqout}{\tilde{X}_{\rm q}} %
\newcommand{\xqbb}{X'_{\rm q}} %

\newcommand{\xkv}{X_{\rm kv}} %
\newcommand{\weit}{T} %
\newcommand{\Nq}{N_{\rm q}} %
\newcommand{\Nkv}{N_{\rm kv}} %
\newcommand{\ques}{Q}
\newcommand{\keys}{K}
\newcommand{\vals}{V}
\newcommand{\posq}{P_{\rm q}} %
\newcommand{\poskv}{P_{\rm kv}} %
\newcommand{\proj}{L}

\newcommand{\dmodel}{d}
\newcommand{\dk}{d'}

\newcommand{\indic}[1]{\mathds{1}_{\{#1\}}}

\newcommand{\bloss}[1]{{\cal L}_{\rm box}(#1)}
\newcommand{\iouloss}[1]{{\cal L}_{\rm iou}(#1)}
\newcommand{\diceloss}[1]{{\cal L}_{\rm DICE}(#1)}
\newcommand{\hloss}[1]{{\cal L}_{\rm Hungarian}(#1)}

\newcommand{\lmatch}[1]{{\cal L}_{\rm match}(#1)}

\def\eqref#1{equation~\ref{#1}}

\def\1{\bm{1}}

\newcommand{\test}{\mathcal{D_{\mathrm{test}}}}

\DeclareMathAlphabet{\mathsfit}{\encodingdefault}{\sfdefault}{m}{sl}
\SetMathAlphabet{\mathsfit}{bold}{\encodingdefault}{\sfdefault}{bx}{n}

\DeclareMathOperator*{\argmin}{arg\,min}

%% file: related.tex
\section{Related work}

Our work build on prior work in several domains: bipartite matching losses for set prediction, encoder-decoder architectures based on the transformer, parallel decoding, and object detection methods.

\subsection{Set Prediction} There is no canonical deep learning model to directly predict sets. The basic set prediction task is multilabel classification (see e.g., \cite{rezatofighi2017deepsetnet,pineda2019ElucidatingIP} for references in the context of computer vision) for which the baseline approach, one-vs-rest, does not apply to problems such as detection where there is an underlying structure between elements (i.e., near-identical boxes). The first difficulty in these tasks is to avoid near-duplicates. Most current detectors use postprocessings such as non-maximal suppression to address this issue, but direct set prediction are postprocessing-free. They need global inference schemes that model interactions between all predicted elements to avoid redundancy. For constant-size set prediction, dense fully connected networks \cite{erhan2014scalable} are sufficient but costly. A general approach is to use auto-regressive sequence models such as recurrent neural networks \cite{vinyals2016order}. In all cases, the loss function should be invariant by a permutation of the predictions. The usual solution is to design a loss  based on the Hungarian algorithm \cite{Kuhn1955TheHM}, to find a bipartite matching between ground-truth and prediction. This enforces permutation-invariance, and guarantees that each target element has a unique match. We follow the bipartite matching loss approach. In contrast to most prior work however, we step away from autoregressive models and use transformers with parallel decoding, which we describe below.

\subsection{Transformers and Parallel Decoding}
Transformers were introduced by Vaswani \etal~\cite{Vaswani2017AttentionIA} as a new attention-based building block for machine translation. Attention mechanisms \cite{bahdanau2015neural} are neural network layers that aggregate information from the entire input sequence. Transformers introduced self-attention layers, which, similarly to Non-Local Neural Networks \cite{Wang2017NonlocalNN}, scan through each element of a sequence and update it by aggregating information from the whole sequence. One of the main advantages of attention-based models is their global computations and perfect memory, which makes them more suitable than RNNs on long sequences. Transformers are now replacing RNNs in many problems in natural language processing, speech processing and computer vision ~\cite{Devlin2019BERTPO,Lscher2019RWTHAS,synnaeve2019end,Radford2019LanguageMA,Parmar2018ImageT}.

Transformers were first used in auto-regressive models, following early sequence-to-sequence models \cite{sutskever2014sequence}, generating output tokens one by one. However, the prohibitive inference cost (proportional to output length, and hard to batch) lead to the development of parallel sequence generation, in the domains of audio \cite{oord2017parallel}, machine translation \cite{Gu2017NonAutoregressiveNM,ghazvininejad2019constant}, word representation learning \cite{Devlin2019BERTPO}, and more recently speech recognition \cite{chan2020imputer}. We also combine transformers and parallel decoding for their suitable trade-off between computational cost and the ability to perform the global computations required for set prediction.

\subsection{Object detection}

Most modern object detection methods make predictions relative to some initial guesses. Two-stage detectors~\cite{Ren2015FasterRT,Cai2019CascadeRH} predict boxes \wrt proposals, whereas single-stage methods make predictions \wrt anchors~\cite{Lin2017FocalLF} or a grid of possible object centers~\cite{Zhou2019objects,tian2019fcos}. Recent work~\cite{Zhang2019bridging} demonstrate that the final performance of these systems heavily depends on the exact way these initial guesses are set. In our model we are able to remove this hand-crafted process and streamline the detection process by directly predicting the set of detections with absolute box prediction \wrt the input image rather than an anchor.

\subsubsection{Set-based loss.} 
Several object detectors~\cite{erhan2014scalable,Liu2016SSDSS,redmon2016you} used the bipartite matching loss. However, in these early deep learning models, the relation between different prediction was modeled with convolutional or fully-connected layers only and a hand-designed NMS post-processing can improve their performance. More recent detectors~\cite{Ren2015FasterRT,Lin2017FocalLF,Zhou2019objects} use non-unique assignment rules between ground truth and predictions together with an NMS.

Learnable NMS methods~\cite{Hosang2017LearningNS,Bodla2017SoftNMSI} and relation networks~\cite{Hu2017RelationNF} explicitly model relations between different predictions with attention. Using direct set losses, they do not require any post-processing steps. However, these methods employ additional hand-crafted context features like proposal box coordinates to model relations between detections efficiently, while we look for solutions that reduce the prior knowledge encoded in the model.

\subsubsection{Recurrent detectors.} 
Closest to our approach are end-to-end set predictions for object detection  \cite{Stewart2015EndtoEndPD} and instance segmentation \cite{RomeraParedes2015RecurrentIS,Park2015LearningTD,ren2017end,Salvador2017RecurrentNN}. Similarly to us, they use bipartite-matching losses with encoder-decoder architectures based on CNN activations to directly produce a set of bounding boxes. These approaches, however, were only evaluated on small datasets and not against modern baselines. In particular, they are based on autoregressive models (more precisely RNNs), so they do not leverage the recent transformers with parallel decoding.

%% file: method.tex
\section{The \detr model}

Two ingredients are essential for direct set predictions in detection: (1) a set prediction loss that %
forces unique matching between predicted and ground truth boxes; (2) an architecture that predicts (in a single pass) a set of objects and models their relation. We describe our architecture in detail in Figure~\ref{fig:architecture}.

\subsection{Object detection set prediction loss}
\label{sec:set_loss}

\detr infers a fixed-size set of $N$ predictions, in a single pass through the decoder, where $N$ is set to be significantly larger than the typical number of objects in an image. One of the main difficulties of training is to score predicted objects (class, position, size) with respect to the ground truth. Our loss produces an optimal bipartite matching between predicted and ground truth objects, and then optimize object-specific (bounding box) losses.

Let us denote by $y$ the ground truth set of objects, and $\hy = \{\hy_i\}_{i=1}^{N}$ the set of $N$ predictions.
Assuming $N$ is larger than the number of objects in the image,
we consider $y$ also as a set of size $N$ padded with $\noobject$ (no object).
To find a bipartite matching between these two sets we search for a permutation of $N$ elements $\sigma \in \Sigma_N$ with the lowest cost:
\begin{equation}
\label{eq:matching}
    \hat{\sigma} = \argmin_{\sigma\in\Sigma_N} \sum_{i}^{N} \lmatch{y_i, \hy_{\sigma(i)}},
\end{equation}
where $\lmatch{y_i, \hy_{\sigma(i)}}$ is a pair-wise \emph{matching cost} between ground truth $y_i$ and a prediction with index $\sigma(i)$. This optimal assignment is computed efficiently with the Hungarian algorithm, following prior work (\eg ~\cite{Stewart2015EndtoEndPD}).

The matching cost takes into account both the class prediction and the similarity of predicted and ground truth boxes. Each element $i$ of the ground truth set can be seen as a $y_i = (c_i, b_i)$ where $c_i$ is the target class label (which may be $\noobject$) and $b_i \in [0, 1]^4$ is a vector that defines ground truth box center coordinates and its height and width relative to the image size. For the prediction with index $\sigma(i)$ we define probability of class $c_i$ as $\hp_{\sigma(i)}(c_i)$ and the predicted box as $\hb_{\sigma(i)}$. With these notations we define
$\lmatch{y_i, \hy_{\sigma(i)}}$ as $-\indic{c_i\neq\noobject}\hp_{\sigma(i)}(c_i) + \indic{c_i\neq\noobject} \bloss{b_{i}, \hb_{\sigma(i)}}$.

This procedure of finding matching plays the same role as the heuristic assignment rules used to match proposal~\cite{Ren2015FasterRT} or anchors~\cite{lin2017feature} to ground truth objects in modern detectors. The main difference is that we need to find one-to-one matching for direct set prediction without duplicates.

The second step is to compute the loss function, the \emph{Hungarian loss} for all pairs matched in the previous step. We define the loss similarly to the losses of common object detectors, \ie a linear combination of a negative log-likelihood for class prediction and a box loss defined later:
\begin{equation}
\hloss{y, \hy} = \sum_{i=1}^N \left[-\log  \hp_{\hat{\sigma}(i)}(c_{i}) + \indic{c_i\neq\noobject} \bloss{b_{i}, \hb_{\hat{\sigma}}(i)}\right]\,,
\end{equation}
where $\hat{\sigma}$ is the optimal assignment computed in the first step (\ref{eq:matching}). In practice, we down-weight the log-probability term when $c_i=\noobject$ by a factor $10$ to account for class imbalance. This is analogous to how Faster R-CNN training procedure balances positive/negative proposals by subsampling \cite{Ren2015FasterRT}.
Notice that the matching cost between an object and $\noobject$ doesn't depend
on the prediction, which means that in that case the cost is a constant.
\oldnew{
There are subtle differences between $\lmatch{\cdot, \cdot}$ and $\hloss{\cdot,
  \cdot}$. Firstly, in}{In} the matching cost we use probabilities
$\hp_{\hat{\sigma}(i)}(c_{i})$ instead of log-probabilities. This makes the
class prediction term commensurable to $\bloss{\cdot, \cdot}$ (described below),
and we observed better empirical performances. \oldnew{The second difference is that in
the $\bloss{\cdot, \cdot}$ described below, we replace the L1 norm by a L2 norm.}{}

\subsubsection{Bounding box loss.} The second part of the matching cost and the Hungarian loss is $\bloss{\cdot}$ that scores the bounding boxes. Unlike many detectors that do box predictions as a $\Delta$ \wrt some initial guesses, we make box predictions directly.
While such approach simplify the implementation it poses an issue  with relative scaling of the loss.
The most commonly-used $\ell_1$ loss will have different scales for small and large boxes even if their relative errors are similar.
To mitigate this issue we use a linear combination of the $\ell_1$ loss and the generalized IoU loss~\cite{Rezatofighi_2018_CVPR} $\iouloss{\cdot, \cdot}$ that is scale-invariant.
Overall, our box loss is $\bloss{b_{i}, \hb_{\sigma(i)}}$ defined as $\lambda_{\rm iou}\iouloss{b_{i}, \hb_{\sigma(i)}} + \lambda_{\rm L1}||b_{i}- \hb_{\sigma(i)}||_1$ where $\lambda_{\rm iou}, \lambda_{\rm L1}\in\Re$ are hyperparameters. 
These two losses are normalized by the number of objects inside the batch.

\subsection{\detr architecture}
\label{sec:architecture}

\begin{figure}[t]
    \centering\small
    \includegraphics[width=1.0\columnwidth]{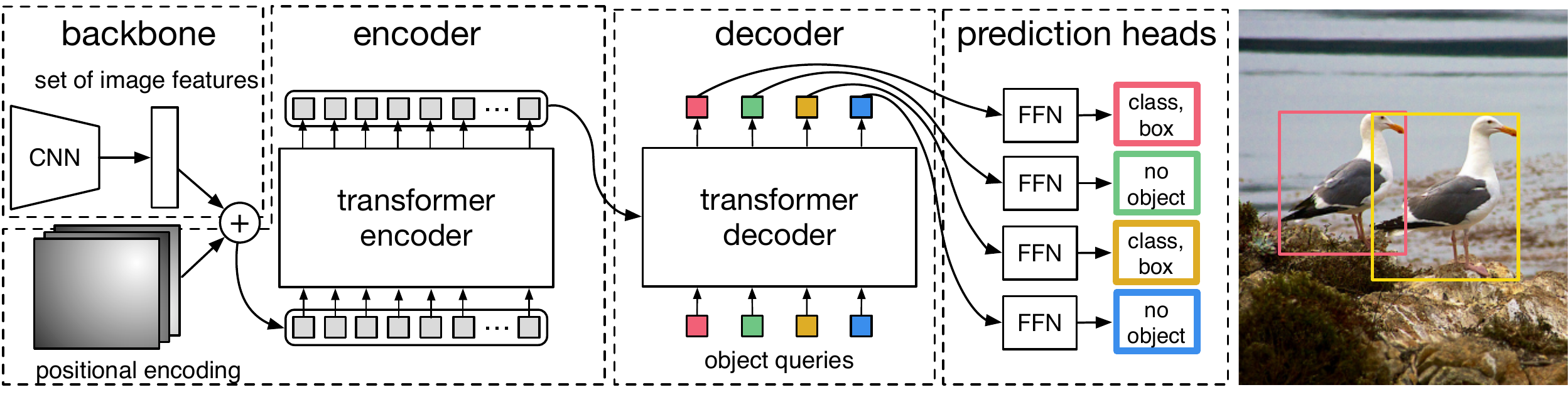}
    \caption{
    \detr uses a conventional CNN backbone to learn a 2D representation of an input image.
    The model flattens it and supplements it with a positional encoding before
    passing it into a transformer encoder.
    A transformer decoder then takes as input a small fixed number of learned positional embeddings,
    which we call \emph{object queries},  and additionally attends to the encoder output.
    We pass each output embedding of the decoder to a shared feed forward
    network (FFN) that predicts either a detection (class and bounding box) or a ``{\tt no object}'' class.
    \label{fig:architecture}
    }
\end{figure}

The overall \detr architecture is surprisingly simple and depicted in  Figure~\ref{fig:architecture}. It contains three main components, which we describe below: a CNN backbone to extract a compact feature representation, an encoder-decoder transformer, and a simple feed forward network~(FFN) that makes the final detection prediction.

Unlike many modern detectors, \detr can be implemented in any deep learning framework that provides a common CNN backbone and a transformer architecture implementation with just a few hundred lines. Inference code for \detr can be implemented in less than 50 lines in PyTorch~\cite{Pytorch}. We hope that the simplicity of our method will attract new researchers to the detection community.

\subsubsection{Backbone.} Starting from the initial image $x_{\rm img} \in\Re^{3\times H_0\times W_0}$ (with $3$ color channels\footnote{The input images are batched together, applying 0-padding adequately to ensure they all have the same dimensions $(H_0,W_0)$ as the largest image of the batch.}), a conventional CNN backbone generates a lower-resolution activation map $f \in\Re^{C\times H\times W}$. Typical values we use are $C=2048$ and $H, W = \frac{H_0}{32}, \frac{W_0}{32}$. 

\subsubsection{Transformer encoder.} First, a 1x1 convolution reduces the channel dimension of the high-level activation map $f$ from $C$ to a smaller dimension $\dmodel$. creating a new feature map $z_0 \in \Re^{\dmodel \times H\times W}$. The encoder expects a sequence as input, hence we collapse the spatial dimensions of $z_0$ into one dimension, resulting in a $\dmodel\times HW$ feature map.
Each encoder layer has a standard architecture and consists of a multi-head self-attention module and a feed forward network (FFN). 
Since the transformer architecture is permutation-invariant, we supplement it %
with fixed positional encodings~\cite{Parmar2018ImageT,Bello2019AttentionAC}
that are added to the input of each attention layer.  We defer to the supplementary material the detailed definition of the architecture, which follows the one described in~\cite{Vaswani2017AttentionIA}.

\subsubsection{Transformer decoder.} 

The decoder follows the standard architecture of the transformer, transforming $N$ embeddings of size $d$ using multi-headed self- and encoder-decoder attention mechanisms. The difference with the original transformer is that our model decodes the $N$ objects in parallel at each decoder layer, while Vaswani et al. \cite{Vaswani2017AttentionIA} use an autoregressive model that predicts the output sequence one element at a time. We refer the reader unfamiliar with the concepts to the supplementary material. %
Since the decoder is also permutation-invariant, the $N$ input embeddings must be different to produce different results. These input embeddings are learnt positional encodings that we refer to as \emph{object queries}, and similarly to the encoder, we add them to the input of each attention layer.
The $N$ object queries are transformed into an output embedding by the decoder. They are then
\emph{independently} decoded into box coordinates and class labels by a %
feed forward network (described in the next subsection), resulting $N$ final predictions. Using self- and encoder-decoder attention over these embeddings, the model globally reasons about all objects %
together using pair-wise relations between them, while being able to use the whole image as context.

\subsubsection{Prediction feed-forward networks (FFNs).}
The final prediction is
computed by \oldnew{two 3-layers}{a 3-layer} perceptron with ReLU activation function and hidden
dimension $d$, \oldnew{}{and a linear projection layer.}
The \oldnew{first}{} FFN predicts the normalized center coordinates, height and width of the box \wrt the input image, and the \oldnew{second one}{linear layer} predicts the class label using a softmax function. Since we predict a fixed-size set of $N$ bounding boxes, where $N$ is usually much larger than the actual number of objects of interest in an image, an additional special class label $\noobject$ is used to represent that no object is detected within a slot. This class plays a similar role to the ``background'' class in the standard object detection approaches.

\subsubsection{Auxiliary decoding losses.}
\label{sec:deep_supervision}
We found helpful to use auxiliary losses~\cite{al2019character} in decoder
during training, especially to help the model output the correct number of
objects of each class.
We add prediction FFNs and Hungarian loss after each decoder layer. \oldnew{The parameters of the prediction FFNs are shared across layers}{All predictions FFNs share their parameters}. We use an additional shared layer-norm to normalize the input to the prediction FFNs from different decoder layers.

\oldnew{\subsection{Extending \detr to higher resolution feature maps}
\label{sec:fpn}

Modern detectors~\cite{tian2019fcos,Zhou2019objects,Du2019SpineNetLS,tan2019efficientdet} adopt feature pyramids~\cite{lin2017feature} with up to stride 8 or 4 features maps and use different levels to detect objects at different scales. Such an approach significantly improves detection performance for small objects. 
We explore an approach to increase the feature resolution popularized by FCIS~\cite{li2017fully}. 
We add a dilation to the last stage of our CNN backbone and remove stride from the first convolution in the stage. This yields a stride 16 feature map instead of the original stride 32. Similarly to previous works we name this design \detr-DC5 (dilated C5 stage). It improves performance for small objects, but increases the computation cost of the self-attentions in encoder by a factor 16. Since encoder self-attentions are only a fraction of the total computation cost, the overall computational cost increases by a factor 2. A full comparison of FLOPs for \detr, \detr-DC5, and Faster R-CNN is given in Table~\ref{table:frcnn}.
}{}

%% file: experiments.tex
\section{Experiments}
\label{sec:experiments}

We show that \detr achieves competitive results compared to Faster R-CNN
in quantitative evaluation on COCO.
Then, we provide a detailed ablation study of the architecture and loss,
with insights and qualitative results.
Finally, to show that \detr is a versatile and extensible model,
we present results on panoptic segmentation, training only a small
extension on a fixed \detr model.
We provide code and pretrained models to reproduce our experiments at
\url{https://github.com/facebookresearch/detr}.
\subsubsection{Dataset.}
We perform experiments on COCO 2017 detection and panoptic segmentation datasets~\cite{Lin2014coco,panoptic_kirillov2019fpn},
containing \oldnew{115k}{118k} training images and 5k validation images.
Each image is annotated with bounding boxes and panoptic segmentation.
There are 7 instances per image on average, up to 63 instances in a single image
in training set, ranging from small to large on the same images. 
If not specified, we report AP as bbox AP, the integral metric over multiple thresholds.
\oldnew{}{For comparison with Faster R-CNN we report validation AP at the last training epoch, for ablations we report median over validation results from the last 10 epochs.}
\subsubsection{Technical details.}
\oldnew{We define a specific instantiation of our model with \detr[-DC5] notation, where $d$ is the dimension of the transformer. Other hyper-parameters are the same for all models and are described in detail in the supplementary.}{}
\oldnew{We train \detr with AdamW~\cite{Loshchilov2017DecoupledWD}, with a weight decay of $10^{-4}$ and initial learning rates of $10^{-4}$ and $10^{-5}$ for the tranformer and the backbone respectively}{We train \detr with AdamW~\cite{Loshchilov2017DecoupledWD}
setting the initial transformer's learning rate to $10^{-4}$, the
backbone's to $10^{-5}$, and weight decay to $10^{-4}$.}
All transformer weights are initialized with Xavier \oldnew{initialization}{init}~\cite{Glorot2010UnderstandingTD},
and the backbone is \oldnew{an}{with} ImageNet-pretrained ResNet model~\cite{He2015DeepRL} from {\sc torchvision} with frozen batchnorm layers.
\oldnew{}{We report results with two different backbones: a ResNet-50 and a ResNet-101. The corresponding models are called respectively \detr and \detr-R101. Following ~\cite{li2017fully}, we also increase the feature resolution by adding a dilation to the last stage of the backbone and removing a stride from the first convolution of this stage. The corresponding models are called respectively \detr-DC5 and \detr-DC5-R101 (dilated C5 stage). This modification increases the resolution by a factor of two, thus improving performance for small objects, at the cost of a 16x higher cost in the self-attentions of the encoder, leading to an overall 2x increase in computational cost. A full comparison of FLOPs of these models and Faster R-CNN is given in Table \ref{table:frcnn}.}

We use scale augmentation, resizing the input images such that the shortest side is at least 480 and at most 800 pixels while the longest at most 1333~\cite{wu2019detectron2}.
To help learning global relationships through the self-attention of the encoder, we also apply random crop augmentations during training, improving the performance by approximately 1 AP.
Specifically, a train image is cropped with probability 0.5 to a
random rectangular patch which is then resized again to 800-1333.
The transformer is trained with default dropout of 0.1.
\oldnew{}{At inference time, some slots predict empty class. To optimize for AP, we override the prediction of these slots with the second highest scoring class, using the corresponding confidence. This improves AP by 2 points compared to filtering out empty slots.}
\oldnew{}{Other training hyperparameters can be found in section~\ref{sec:hyperparameters}.}
For our ablation experiments we use training schedule 
of 300 epochs with a learning rate drop by a factor of 10 after 200 epochs,
where a single epoch is a pass over all training images once.
\oldnew{Training the baseline model for 300 epochs on 128 V100 GPUs takes about 24 hours, with one image per GPU (hence a total batch size of 128).}
{Training the baseline model for 300 epochs on 16 V100 GPUs takes 3 days,
with 4 images per GPU (hence a total batch size of 64).}
For the longer schedule used to compare with Faster R-CNN we train
\oldnew{for 900 epochs with learning rate drops by 2 every 50 epochs starting from epoch 600. This schedule adds 3 AP compared to the short 300 epoch schedule.}
{for 500 epochs with learning rate drop after 400 epochs.
This schedule adds 1.5 AP compared to the shorter schedule.}

\subsection{Comparison with Faster R-CNN}

\input{table1_alt}

Transformers are typically trained with Adam or Adagrad optimizers with
very long training schedules and dropout, and this is true for \detr as well.
Faster R-CNN, however, is trained with SGD with minimal data augmentation
and we are not aware of successful applications of Adam or dropout.
Despite these differences we attempt to make a Faster R-CNN baseline stronger.
To align it with \detr,
we add generalized IoU~\cite{Rezatofighi_2018_CVPR} to the box loss, the same random crop augmentation and long training known to
improve results~\cite{He2018RethinkingIP}.
Results are presented in Table~\ref{table:frcnn}.
In the top section we show Faster R-CNN results from Detectron2 Model Zoo~\cite{wu2019detectron2}
for models trained with the \texttt{3x} schedule.
In the middle section we show results (with a ``+'') for the same models
but trained with the \texttt{9x} schedule (109 epochs) and the described enhancements,
which in total adds 1-2 AP. In the last section of Table~\ref{table:frcnn}
we show the results for multiple \detr models.
To be comparable in the number of parameters we choose a model
with 6 transformer and 6 decoder layers of width 256\oldnew{}{ with 8 attention heads}.
Like Faster R-CNN with FPN this model has 41.3M parameters,
out of which 23.5M are in ResNet-50, and 17.8M are in the transformer.
Even though both Faster R-CNN and \detr are still likely to
further improve with longer training,
we can conclude that \oldnew{\detr-DC5}{\detr} can be competitive with Faster R-CNN
with the same number of parameters, achieving 42 AP on the COCO val subset.
\oldnew{The way \detr achieves this is by improving \AP{L}, however note that the model is still lagging behind in \AP{S}.}
{The way \detr achieves this is by improving \AP{L} (+7.8),
however note that the model is still lagging behind in \AP{S} (-5.5).
\detr-DC5 with the same number of parameters and similar FLOP count has higher AP,
but is still significantly behind in \AP{S} too.}
\oldnew{}{Faster R-CNN and \detr with ResNet-101 backbone show comparable results as well.}

\subsection{Ablations}

Attention mechanisms in the transformer decoder are the key components which model relations between feature representations of different detections. In our ablation analysis, we explore how other components of our architecture and loss influence the final performance. For the study we choose ResNet-50-based \detr model with 6 encoder, 6 decoder layers and width 256.
The model has 41.3M parameters,
achieves \oldnew{37.3 and 40.3 AP}{40.6 and 42.0 AP} on short and long schedules respectively,
and runs at \oldnew{23}{28} FPS, similarly to Faster R-CNN-FPN with the same backbone.
\subsubsection{Number of encoder layers.}
\begin{figure}[t]
    \centering\small
    \includegraphics[width=1.0\columnwidth]{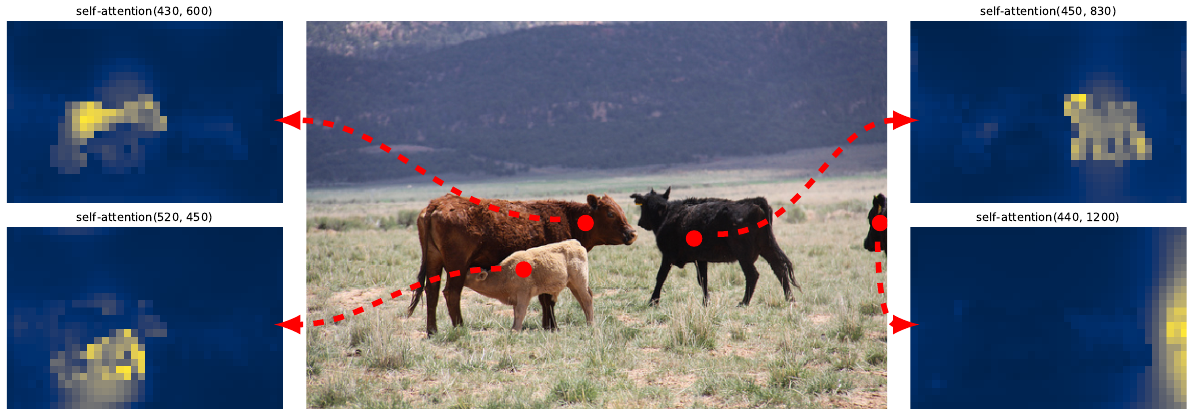}
    \caption{%
    Encoder self-attention for a set of reference points. The encoder is able to separate individual instances.
    Predictions are made with baseline \detr model on a validation set image.
    }
    \label{fig:encoder_self_attention}
\end{figure}

We evaluate the importance of global image-level self-attention by changing the number of encoder layers (Table~\ref{table:enc_layers}).
Without encoder layers, overall AP drops by \oldnew{3.4}{3.9} points, with a more significant drop of \oldnew{5.5}{6.0 AP} on large objects.
We hypothesize that, by using global scene reasoning, the encoder is important for disentangling objects.
In Figure~\ref{fig:encoder_self_attention}, we visualize the attention maps of the last encoder layer of a trained model, focusing on a few points in the image.
The encoder seems to separate instances already, which likely simplifies object extraction and localization for the decoder.
\setlength{\tabcolsep}{0.78em}
\begin{table}[t]
    \centering\small
    \caption{%
    Effect of encoder size. Each row corresponds to a model with varied number of encoder layers and fixed number of decoder layers.
    Performance gradually improves with more encoder layers.\oldnew{, mostly due to large and medium objects}{}}
    \begin{tabular}{lccccccc}
        \toprule
        \#layers & GFLOPS/FPS & \#params & AP & \AP{50} & \AP{S} & \AP{M} & \AP{L} \\
        \midrule
        0 & 76/28 & 33.4M & 36.7 & 57.4 & 16.8 & 39.6 & 54.2 \\
        3 & 81/25 & 37.4M & 40.1 & 60.6 & 18.5 & 43.8 & 58.6 \\
        6 & 86/23 & 41.3M & 40.6 & 61.6 & 19.9 & 44.3 & 60.2 \\
        12 & 95/20 & 49.2M & 41.6 & 62.1 & 19.8 & 44.9 & 61.9 \\
        \bottomrule
    \end{tabular}
    \label{table:enc_layers}
\end{table}
\subsubsection{Number of decoder layers.}
\begin{figure}[t]
    \centering\small
    \begin{minipage}[t!]{0.57\textwidth}
        \centering\small
        \includegraphics[width=0.85\columnwidth]{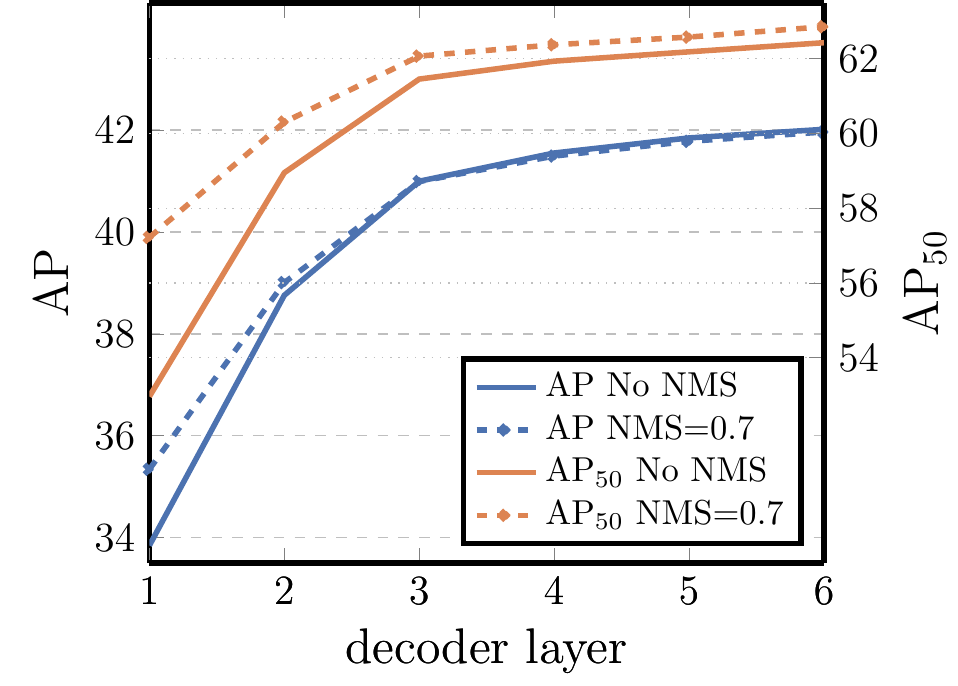}
        \caption{%
        AP and \AP{50} performance after each decoder layer.
        A single long schedule baseline model is evaluated.
        \detr does not need NMS by design, which is validated by this figure.
        NMS lowers AP in the final layers, removing TP predictions,
        but improves AP in the first decoder layers, removing double predictions,
        as there is no communication in the first layer\oldnew{}{, and slightly improves \AP{50}.}
        }
        \label{fig:nms}
    \end{minipage}
    \hskip1em
    \begin{minipage}[t!]{0.35\textwidth}
        \centering\small
        \vskip0.8em
        \includegraphics[width=\columnwidth]{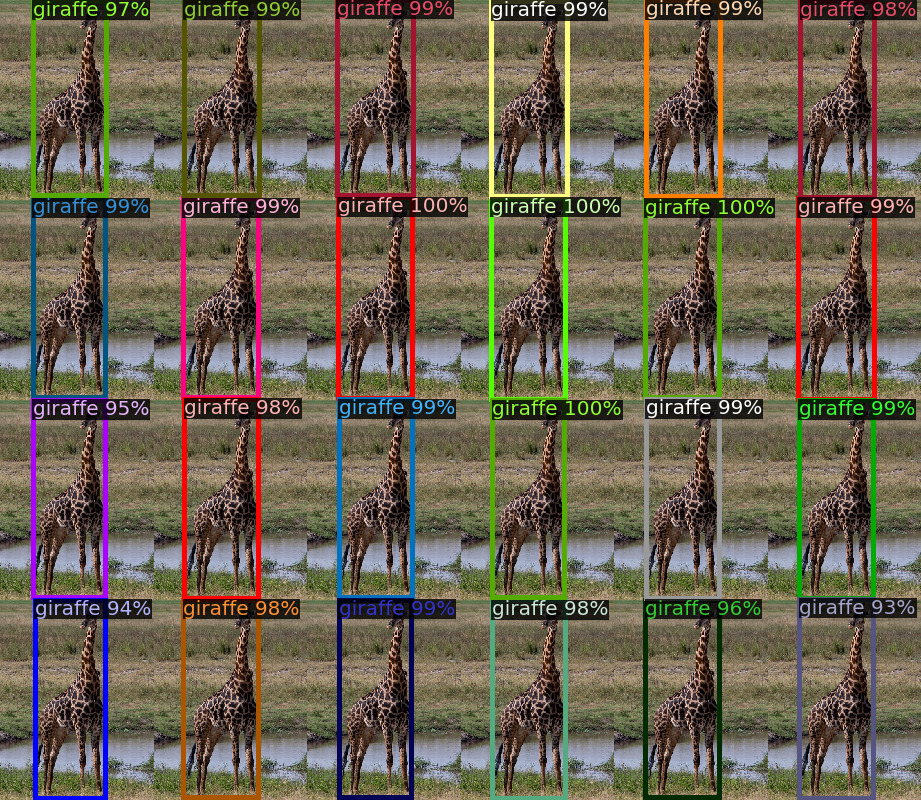}
        \vskip0.8em
        \caption{%
        Out of distribution generalization for rare classes.
        Even though no image in the training set has more than 13 giraffes, 
        \detr has no difficulty generalizing to 24 and more instances of the same class.
        }
        \label{fig:giraffe}
    \end{minipage}
\end{figure}
We apply auxiliary losses after each decoding layer
(see Section \ref{sec:deep_supervision}), hence, the prediction FFNs are trained by design to predict objects out of the outputs of every decoder layer. 
We analyze the importance of each decoder layer by evaluating the objects that would be predicted at each stage of the decoding (Fig.~\ref{fig:nms}).
Both AP and \AP{50} improve after every layer, totalling into a  very significant \oldnew{+5.4/6.3}{+8.2/9.5} AP improvement between the first and the last layer.\oldnew{even though we observe some
stagnation after layer 5.}{}
With its set-based loss, \detr does not need NMS by design. To verify this we run a standard NMS procedure with default parameters~\cite{wu2019detectron2} for the outputs after each decoder.
NMS  improves performance for the predictions from the first decoder. This can be explained by the fact that a single decoding layer of the transformer is not able to compute any cross-correlations between the output elements, and thus it is prone to making multiple predictions for the same object.
In the second and subsequent layers, the self-attention mechanism over the activations allows the model to inhibit duplicate predictions. We observe that the improvement brought by NMS diminishes as depth increases. At the last layers, we observe a small loss in AP as NMS incorrectly removes true positive predictions.

\begin{figure}[t]
    \centering\small
    \includegraphics[height=0.43\columnwidth]{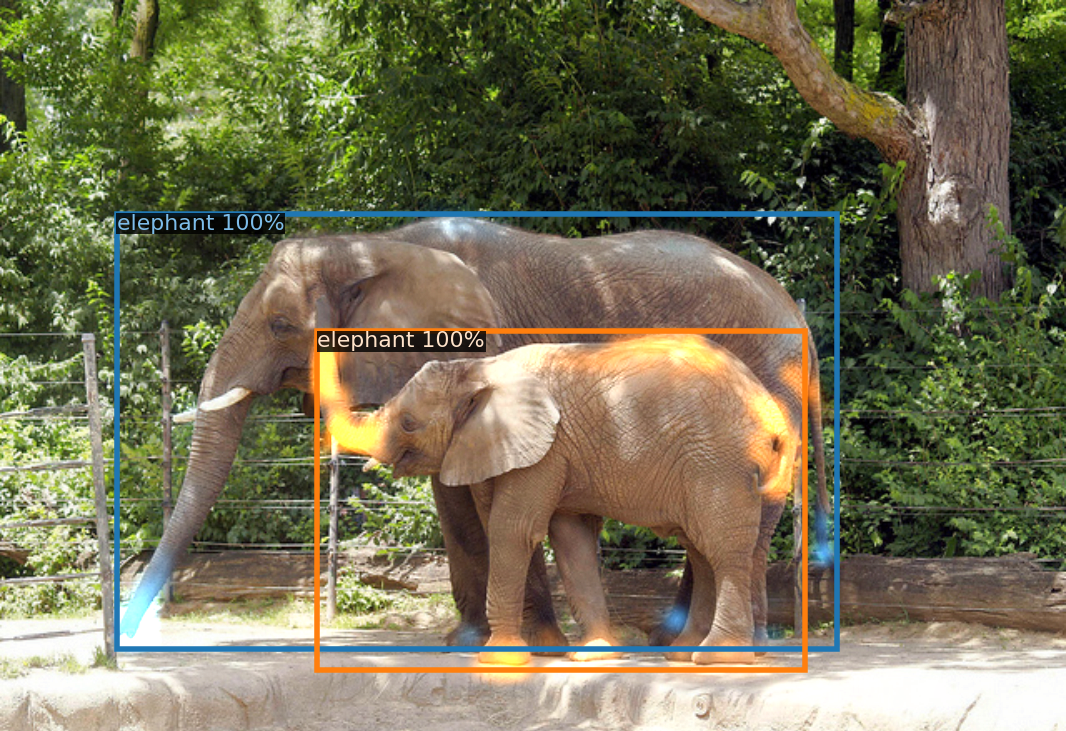}
    \includegraphics[height=0.43\columnwidth]{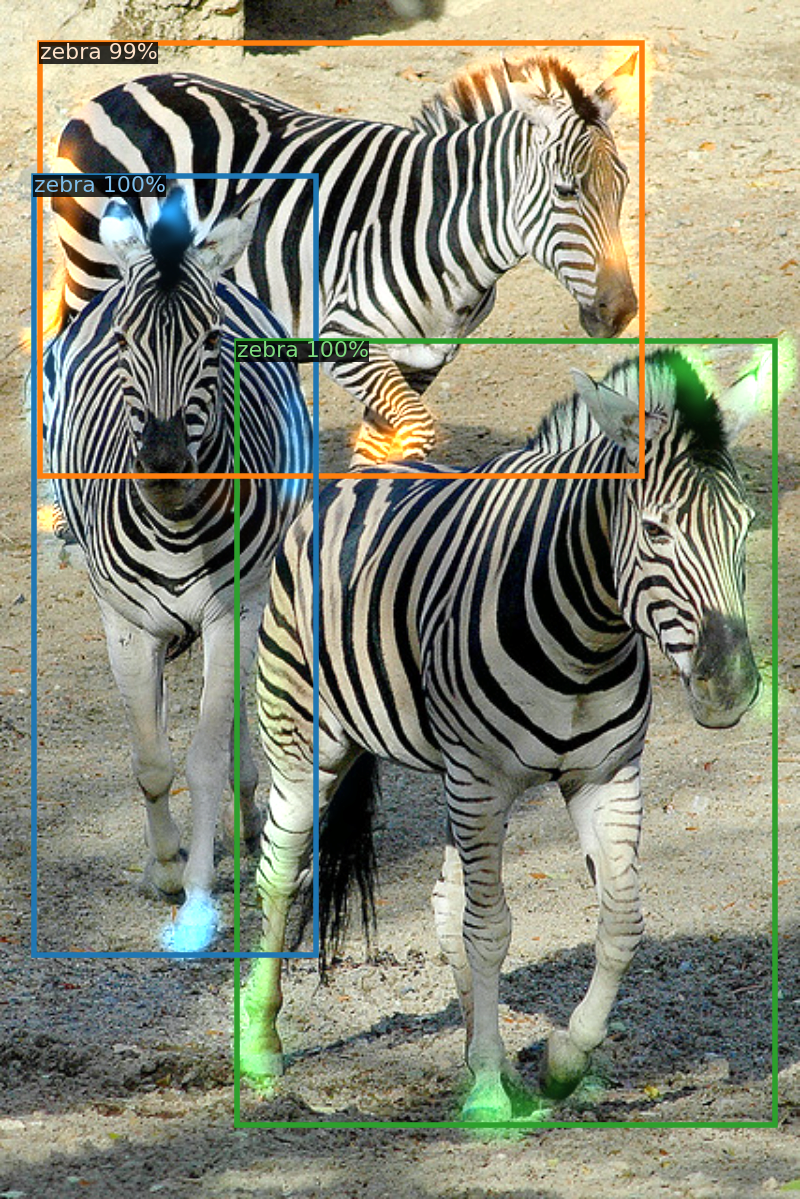}
    \caption{%
    Visualizing decoder attention for every predicted object
    (images from COCO \texttt{val} set).
    Predictions are made with \detr-DC5 model.
    Attention scores are coded with different colors for different objects.
    Decoder typically attends to object extremities,
    such as legs and heads.
    Best viewed in color.
    }
    \label{fig:attention_maps}
\end{figure}

Similarly to visualizing encoder attention,
we visualize decoder attentions in Fig.~\ref{fig:attention_maps},
coloring attention maps for each predicted object in different colors.
We observe that decoder attention is fairly local, meaning that it mostly attends to object extremities such as heads or legs.
We hypothesise that after the encoder has separated instances via
global attention, the decoder only needs to attend to the extremities to extract the class and object boundaries.

\subsubsection{Importance of FFN.}
FFN inside tranformers can be seen as $1\times1$ convolutional layers,
making encoder similar to
attention augmented convolutional networks~\cite{Bello2019AttentionAC}.
We attempt to remove it completely leaving only attention in the transformer layers.
By reducing the number of network parameters from 41.3M to 28.7M,
leaving only 10.8M in the transformer, performance drops by \oldnew{almost 2 AP}{2.3 AP}, we thus conclude that FFN are important for achieving good results.

\subsubsection{Importance of positional encodings.}
There are two kinds of positional encodings in our model: spatial 
positional encodings and output positional encodings (object queries).
We experiment with various combinations of fixed and learned encodings,
results can be found in table~\ref{table:pos_enc}.
Output positional encodings are required and cannot be removed,
so we experiment with either passing them once at decoder input
or adding to queries at every decoder attention layer.
In the first experiment we completely remove spatial positional encodings
and pass output positional encodings at input
and, interestingly, the model still achieves 
more than \oldnew{31}{32} AP, losing \oldnew{6}{7.8} AP to the baseline.
Then, we pass fixed sine spatial positional encodings and the output encodings
at input once, as in the original transformer~\cite{Vaswani2017AttentionIA},
and find that this leads to \oldnew{1.1}{1.4} AP drop compared to passing the positional encodings
directly in attention.
Learned spatial encodings passed to the attentions give
similar results.
Surprisingly, we find that not passing any spatial encodings in the encoder
only leads to a minor AP drop of 1.3 AP.
When we pass the encodings to the attentions, they are shared across all layers,
and the output encodings (object queries) are always learned.
\setlength{\tabcolsep}{6pt}
\begin{table}[h!]
    \centering\small
    \caption{Results for different positional encodings compared
    to the baseline (last row), which has fixed sine pos. encodings passed
    at every attention layer in both the encoder and the decoder.
    Learned embeddings are shared between all layers.
    Not using spatial positional encodings leads to a significant
    drop in AP. Interestingly, passing them in decoder only
    leads to a minor AP drop.
    All these models use learned output positional encodings.
    \label{table:pos_enc}}
    \begin{tabular}{lll|cc|cc}
        \toprule
        \multicolumn{2}{c}{spatial pos. enc.} & output pos. enc.\\
        \multicolumn{1}{c}{encoder} & \multicolumn{1}{c}{decoder} & \multicolumn{1}{c|}{decoder} & AP & $\Delta$ & \AP{50} & $\Delta$ \\
        \midrule
        none                & none & learned at input
        & 32.8 & -7.8 & 55.2 & -6.5 \\

        sine at input        & sine at input & learned at input
        & 39.2 & -1.4 & 60.0 & -1.6 \\

        learned at attn. & learned at attn. & learned at attn.
        & 39.6 & -1.0 & 60.7 & -0.9 \\

        none & sine at attn. & learned at attn.
        & 39.3 & -1.3 & 60.3 & -1.4 \\
        
        sine at attn. & sine at attn. & learned at attn.
        & \textbf{40.6} & - & \textbf{61.6} & - \\
        \bottomrule
    \end{tabular}
\end{table}

Given these ablations, we conclude that transformer components:
the global self-attention in encoder, FFN, multiple decoder layers, and positional encodings,
all significantly contribute to the final object detection performance.

\subsubsection{\oldnew{Cost}{Loss} ablations.} 
To evaluate the importance of different components of the matching cost and the loss,
we train several models turning them on and off.
There are three components to the loss: \oldnew{classification cost, bounding box distance cost}
{classification loss, $\ell_1$ bounding box distance loss,} and GIoU~\cite{Rezatofighi_2018_CVPR} loss.
The classification \oldnew{cost}{loss} is essential for training and cannot be turned off,
so we train a model without bounding box distance loss,
and a model without the GIoU loss,
and compare with baseline, trained with all three losses.
Results are presented in table~\ref{table:cost}.
\setlength{\tabcolsep}{0.78em}
\begin{table}[h!]
    \centering\small
    \caption{Effect of \oldnew{cost}{loss} components on AP.
    \oldnew{We train two models turning off box cost and GIoU cost,
    and observe that box cost gives better results on large objects,
    whereas GIoU cost is better on small.}
    {We train two models turning off $\ell_1$ loss, and GIoU loss,
    and observe that $\ell_1$ gives poor results on its own,
    but when combined with GIoU improves \AP{M} and \AP{L}.
    Our baseline (last row) combines both losses.}
    }
    \begin{tabular}{ccc|cc|cc|ccc}
        \toprule
        class & $\ell_1$ & GIoU & AP & $\Delta$ & \AP{50} & $\Delta$ & \AP{S} & \AP{M} & \AP{L} \\
        \midrule
        \chk & \chk & &
        35.8 & -4.8 & 57.3 & -4.4 & 13.7 &	39.8 &	57.9 \\
        \chk & & \chk &
        39.9 & -0.7 & \textbf{61.6} & 0 &	\textbf{19.9} &	43.2 &  57.9 \\
        \chk & \chk & \chk &
        \textbf{40.6} & - & \textbf{61.6} & - & \textbf{19.9} &	\textbf{44.3} &	\textbf{60.2} \\
        \bottomrule
    \end{tabular}
    \label{table:cost}
\end{table}
\oldnew{The GIoU cost shows better results on small objects (+2.9 AP difference),
whereas the bounding box cost improves results on large objects (+5.5 AP difference),
higher than the baseline.}
{GIoU loss on its own accounts for most of the model performance,
losing only 0.7 AP to the baseline with combined losses.
Using $\ell_1$ without GIoU shows poor results.}
We only studied simple ablations of different \oldnew{costs}{losses} (using the same weighting every time),
but other means of combining them may achieve different results.

\subsection{Analysis}
\subsubsection{Decoder output slot analysis}

\begin{figure}[t]
    \centering
    \includegraphics[width=\columnwidth]{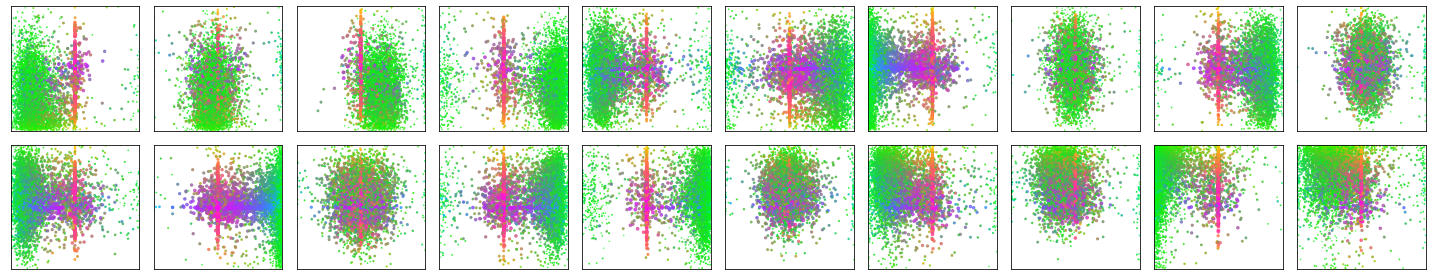}
    \caption{
    Visualization of all box predictions on all images from COCO 2017 val set for 20 out of total $N=100$ prediction slots in \detr decoder. Each box prediction is represented as a point with the coordinates of its center in the 1-by-1 square normalized by each image size. The points are color-coded so that green color corresponds to small boxes, red to large horizontal boxes and blue to large vertical boxes. We observe that each slot learns to specialize on certain areas and box sizes with several operating modes. We note that almost all slots have a mode of predicting large image-wide boxes that are common in COCO dataset.
    }
    \label{fig:queries}
\end{figure}

In Fig.~\ref{fig:queries} we visualize the boxes predicted by different slots for
all images in COCO 2017 val set. \detr learns different specialization for each
query slot. We observe that each slot has several modes of operation focusing on
different areas and box sizes. In particular, all slots have the mode for
predicting image-wide boxes (visible as the red dots aligned in the middle of
the plot). We hypothesize that this is related to the distribution of objects in COCO.
\subsubsection{Generalization to unseen numbers of instances.}
Some classes in COCO are not well represented with many instances of the same class in the same image. For example, there is no image with more than 13 giraffes in the training set. We create a synthetic image\footnote{Base picture credit: https://www.piqsels.com/en/public-domain-photo-jzlwu} to verify the generalization ability of \detr (see Figure~\ref{fig:giraffe}). Our model is able to find all 24 giraffes on the image which is clearly out of distribution. This experiment confirms that there is no strong class-specialization in each object query.

\subsection{\detr for panoptic segmentation}

\begin{figure}[h]
    \centering\small
    \includegraphics[width=1.0\columnwidth]{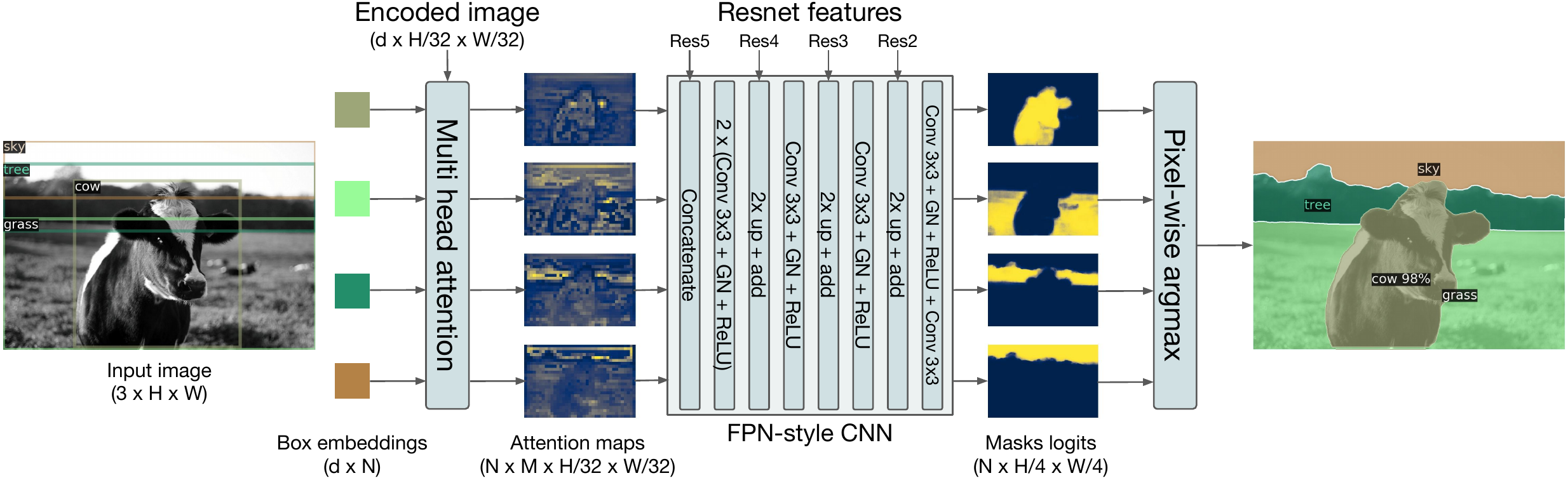}
    \caption{
    Illustration of the panoptic head. A binary mask is generated in parallel for each detected object, then the masks are merged using  pixel-wise argmax. %
    }
    \label{fig:pano}
\end{figure}
\begin{figure}[h]
    \centering\small
    \includegraphics[height=0.27\columnwidth]{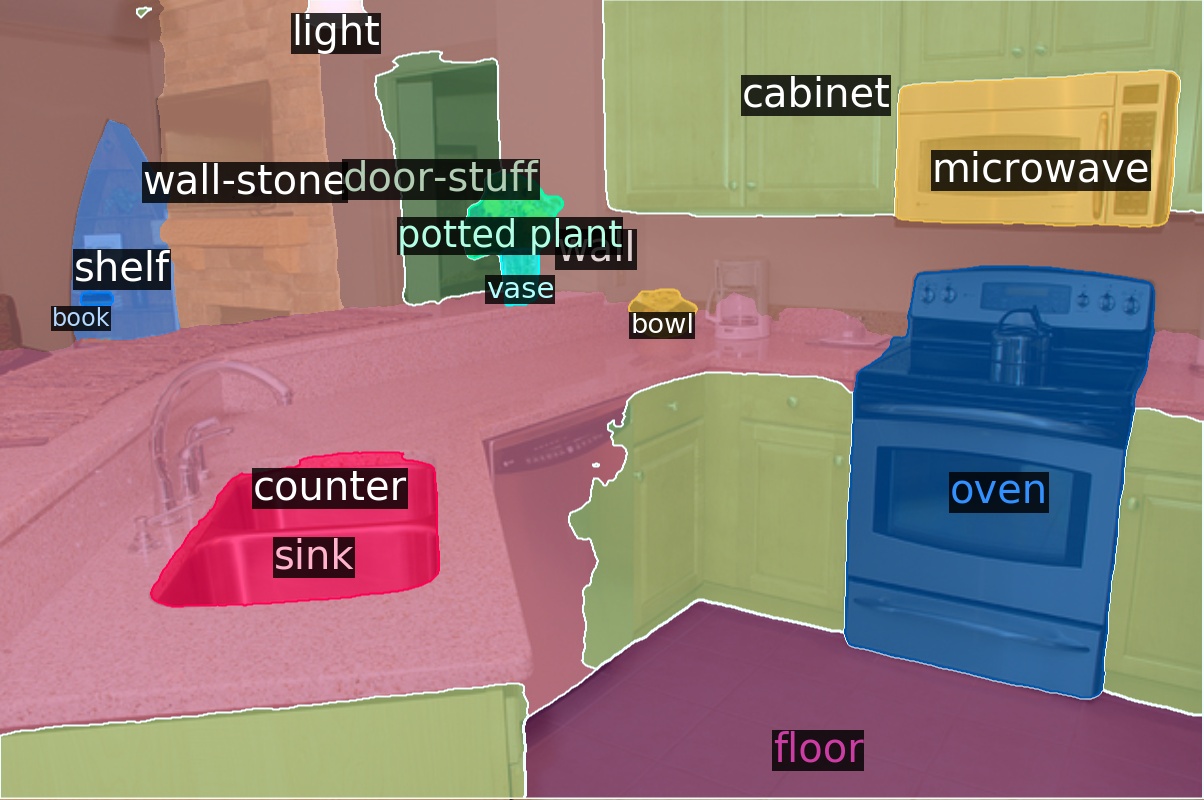}
    \includegraphics[height=0.27\columnwidth]{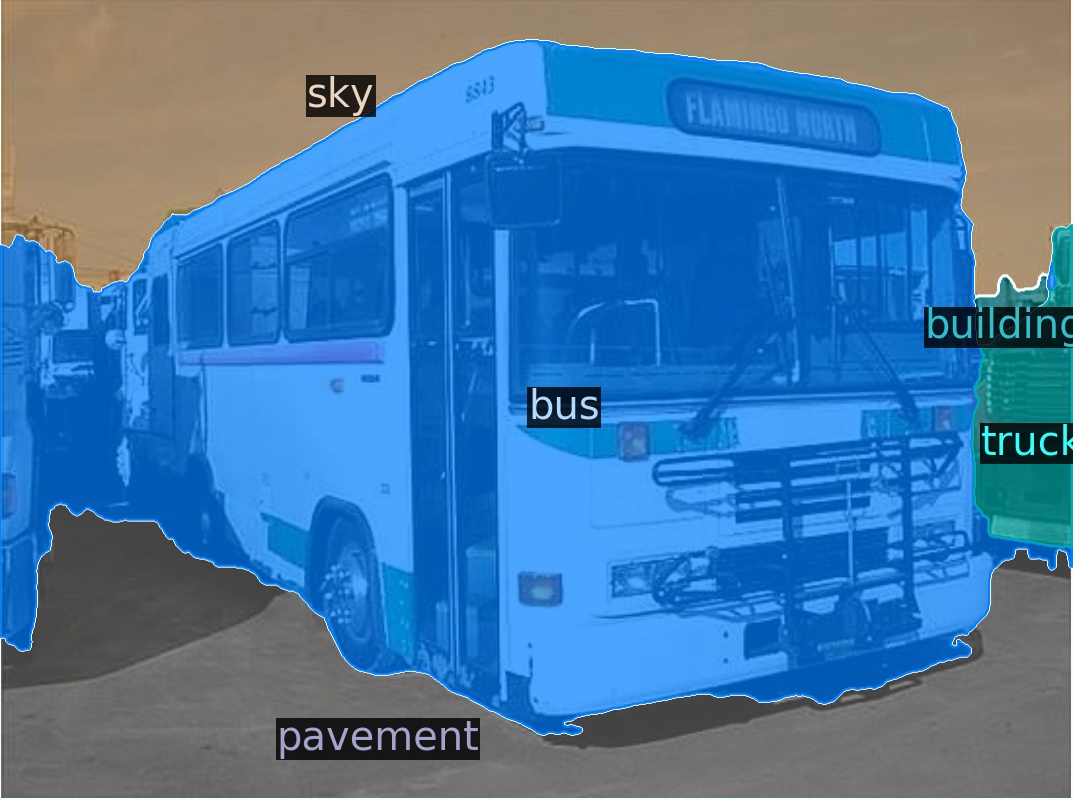}
    \includegraphics[height=0.27\columnwidth]{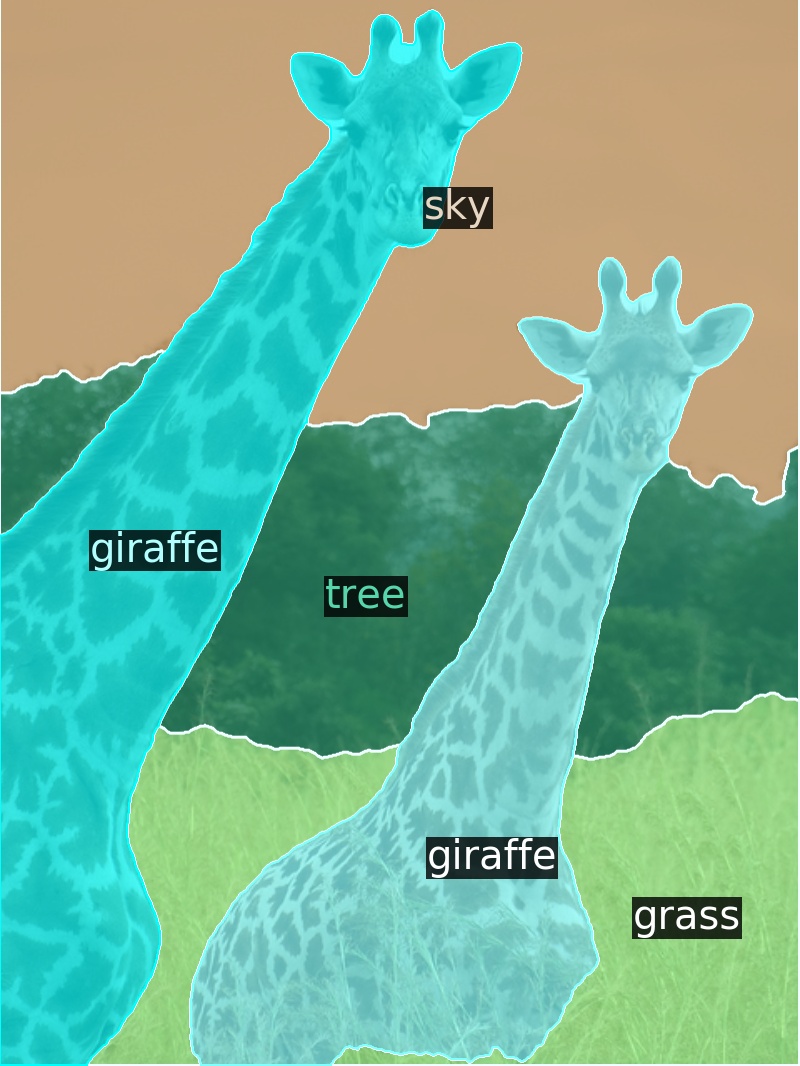}
    \caption{
    Qualitative results for panoptic segmentation generated by DETR-R101. \detr produces aligned mask predictions in a unified manner for things and stuff.
    }
    \label{fig:pano_quali}
\end{figure}

Panoptic segmentation~\cite{Kirillov2019panoptic} has recently attracted a lot of attention from the computer vision community. Similarly to the extension of Faster R-CNN~\cite{Ren2015FasterRT} to Mask R-CNN~\cite{He2017MaskR}, \detr can be naturally extended by adding a mask head on top of the decoder outputs. In this section we demonstrate that such a head can be used to produce panoptic segmentation~\cite{Kirillov2019panoptic} by treating stuff and thing classes in a unified way. We perform our experiments on the panoptic annotations of the COCO dataset that has 53 stuff categories in addition to 80 things categories.

We train \detr to predict boxes around both \emph{stuff} and \emph{things}
classes on COCO, using the same recipe. Predicting boxes is required for the
training to be possible, since the Hungarian matching is computed using
distances between boxes. We also add a mask head which predicts a binary mask
for each of the predicted boxes, see Figure~\ref{fig:pano}. It takes as input
the output of transformer decoder for each object and computes multi-head (with
$M$ heads) attention scores of this embedding over the output of the encoder,
generating $M$ attention heatmaps per object in a small resolution. To make the
final prediction and increase the resolution, an FPN-like architecture is used.
We describe the architecture in more details in the supplement. The final
resolution of the masks has stride 4 and each mask is supervised independently
using the DICE/F-1 loss~\cite{milletari2016v} \oldnew{}{and Focal loss~\cite{Lin2017FocalLF}}.

The mask head can be trained either jointly, or in a two steps \oldnew{}{process}, where we
train \detr for boxes only, then freeze all the weights and train only the mask
head for 25 epochs. Experimentally, these two approaches give similar results,
we report results using the latter method since it results in a shorter total
wall-clock time training.

To predict the final panoptic segmentation we simply use an argmax over the mask scores at each pixel, and assign the corresponding categories to the resulting masks. This procedure guarantees that the final masks have no overlaps and, therefore, \detr does not require a heuristic~\cite{Kirillov2019panoptic} that is often used %
to align different masks.

\subsubsection{Training details.} We train \detr, \detr-DC5 and \detr-R101 models
following the recipe for bounding box detection to predict boxes around stuff
and things classes in COCO dataset.  
\oldnew{During inference we collapse different mask predictions of the same stuff
category in one. The new mask head is trained for 25 epochs (see supplementary
for details). Similarly to~\cite{Kirillov2019panoptic}, we remove small stuff
(resp. things) predictions that have area smaller than 256 pixels (resp 4
pixels) as they are likely to be spurious segments, and keep only the detections
with a confidence higher than 75\%.}{The new mask head is trained for 25 epochs (see supplementary
for details). During inference we first filter out the detection with a confidence below 85\%, then compute the per-pixel argmax to determine in which mask each pixel belongs. We then collapse different mask predictions of the same stuff
category in one, and filter the empty ones (less than 4 pixels).}

\setlength{\tabcolsep}{0.3em}
\begin{table}[t]
  \centering\small
  \caption{%
    Comparison with the state-of-the-art methods
    UPSNet~\cite{panoptic_xiong2019upsnet} and Panoptic
    FPN~\cite{panoptic_kirillov2019fpn} on the COCO \texttt{val} dataset
    We retrained PanopticFPN with the same data-augmentation as DETR,
    on a 18x schedule for fair comparison. UPSNet uses the \texttt{1x} schedule,
    UPSNet-M is the version with multiscale test-time augmentations.}
  \resizebox{\columnwidth}{!}{
  \begin{tabular}{lc|ccc|ccc|ccc|c}
    \toprule
    Model & Backbone & PQ & SQ & RQ & $\text{PQ}^\text{th}$ & $\text{SQ}^\text{th}$  & $\text{RQ}^\text{th}$  & $\text{PQ}^\text{st}$ & $\text{SQ}^\text{st}$ & $\text{RQ}^\text{st}$ & AP \\
    
    \midrule
    PanopticFPN++  & R50 & 42.4 & 79.3 & 51.6 & 49.2 & 82.4 & 58.8 & 32.3 & 74.8 & 40.6 & 37.7\\
    UPSnet & R50  & 42.5 & 78.0 & 52.5 & 48.6 & 79.4 & 59.6 & 33.4 & 75.9 & 41.7 & 34.3 \\
    UPSnet-M & R50  & 
                     43.0  & 79.1 &  52.8 & 48.9 & 79.7 & 59.7 & 34.1  & 78.2 & 42.3 & 34.3 \\
    PanopticFPN++ & R101 & 44.1 & 79.5 & 53.3 & \textbf{51.0} & \textbf{83.2} & 60.6 & 33.6 & 74.0 & 42.1 & \textbf{39.7} \\

    DETR & R50 & 43.4  & 79.3 &  53.8 &  48.2 &  79.8  & 59.5 & 36.3  & 78.5  & 45.3 & 31.1 \\
    DETR-DC5 & R50& 44.6 & 79.8 & 55.0 & 49.4 & 80.5 & 60.6 & \textbf{37.3} & \textbf{78.7} & \textbf{46.5} & 31.9\\
    DETR-R101 & R101& \textbf{45.1} & \textbf{79.9} & \textbf{55.5} & 50.5 & 80.9 & \textbf{61.7} & 37.0 & 78.5 & 46.0 & 33.0\\
    \bottomrule
  \end{tabular}
}
  \label{table:panoptic_numbers}
\end{table}

\subsubsection{Main results.} Qualitative results are shown in
Figure~\ref{fig:pano_quali}. In table~\ref{table:panoptic_numbers}  we compare
our unified panoptic segmenation approach with several established methods that
treat things and stuff differently. We report the Panoptic Quality (PQ) and the
break-down on things ($\text{PQ}^\text{th}$) and stuff ($\text{PQ}^\text{st}$). We also report the mask AP (computed on the
things classes), before any panoptic post-treatment (in our case, before taking
the pixel-wise argmax).
We show that \detr outperforms published results on COCO-val 2017, as well as
our strong PanopticFPN baseline (trained with same data-augmentation 
as \detr, for fair comparison). The result break-down shows that \detr is
especially dominant on stuff classes, and we hypothesize that the global
reasoning allowed by the encoder attention is the key element to this result.
For things class, despite a severe deficit of up to 8 mAP compared to the
baselines on the mask AP computation, \detr obtains competitive
$\text{PQ}^\text{th}$.
\oldnew{}{We also evaluated our method on the test set of the COCO dataset, and obtained 46 PQ.}
We hope that our approach will inspire the exploration of fully unified models for panoptic segmentation in future work.

%% file: table1_alt.tex
\begin{table}[t]
    \centering\small
    \caption{Comparison with Faster R-CNN with a ResNet-50 and ResNet-101 backbones on the COCO validation set.
    The top section shows results for Faster R-CNN models in Detectron2~\cite{wu2019detectron2},
    the middle section shows results for Faster R-CNN models
    with GIoU~\cite{Rezatofighi_2018_CVPR}, random crops train-time augmentation, and the long \texttt{9x} training schedule.
    \detr models achieve comparable results
    to heavily tuned Faster R-CNN baselines, having lower \AP{S} but greatly improved \AP{L}.
    \oldnew{}{We use torchscript Faster R-CNN and \detr models to measure FLOPS and FPS. Results without R101 in the name correspond to ResNet-50.}
    }
    \begin{tabular}{lcccccccc}
        \toprule
        Model & GFLOPS/FPS & \#params & AP & \AP{50} & \AP{75} & \AP{S} & \AP{M} & \AP{L} \\
        \midrule
        Faster RCNN-DC5 & 320/16 & 166M  & 39.0 & 60.5 & 42.3 & 21.4 & 43.5 & 52.5 \\
        Faster RCNN-FPN & 180/26 & 42M & 40.2 & 61.0 & 43.8 & 24.2 & 43.5 & 52.0 \\
        Faster RCNN-R101-FPN & 246/20 & 60M & 42.0 & 62.5 & 45.9 & 25.2 & 45.6 & 54.6 \\
        \midrule
        Faster RCNN-DC5+ & 320/16 & 166M & 41.1 & 61.4 & 44.3 & 22.9 & 45.9 & 55.0\\
        Faster RCNN-FPN+ & 180/26 & 42M & 42.0 & 62.1 & 45.5 & 26.6 & 45.4 & 53.4\\
        Faster RCNN-R101-FPN+ & 246/20 & 60M & 44.0 & 63.9 & \textbf{47.8} & \textbf{27.2} & 48.1 & 56.0\\
        \midrule
        {\detr} & 86/28 & 41M & 42.0  & 62.4  & 44.2  & 20.5  & 45.8  & 61.1 \\
        {\detr}-DC5 & 187/12 & 41M & 43.3  & 63.1  & 45.9  & 22.5  & 47.3  & 61.1 \\
        {\detr}-R101 & 152/20 & 60M & 43.5  & 63.8  & 46.4  & 21.9  & 48.0  & 61.8 \\
        \detr-DC5-R101 & 253/10 & 60M & \textbf{44.9}  & \textbf{64.7}  & 47.7  & 23.7  & \textbf{49.5}  & \textbf{62.3} \\
        \bottomrule
    \end{tabular}
    \label{table:frcnn}
\end{table}

%% file: supplementary.tex
\section{Appendix}

\input{preliminaries}

\subsection{Detailed architecture}\label{sec:detailed_arch}
\begin{figure}
    \centering\small
    \includegraphics[clip, trim=0cm 0cm 0cm 0cm, width=0.75\columnwidth]{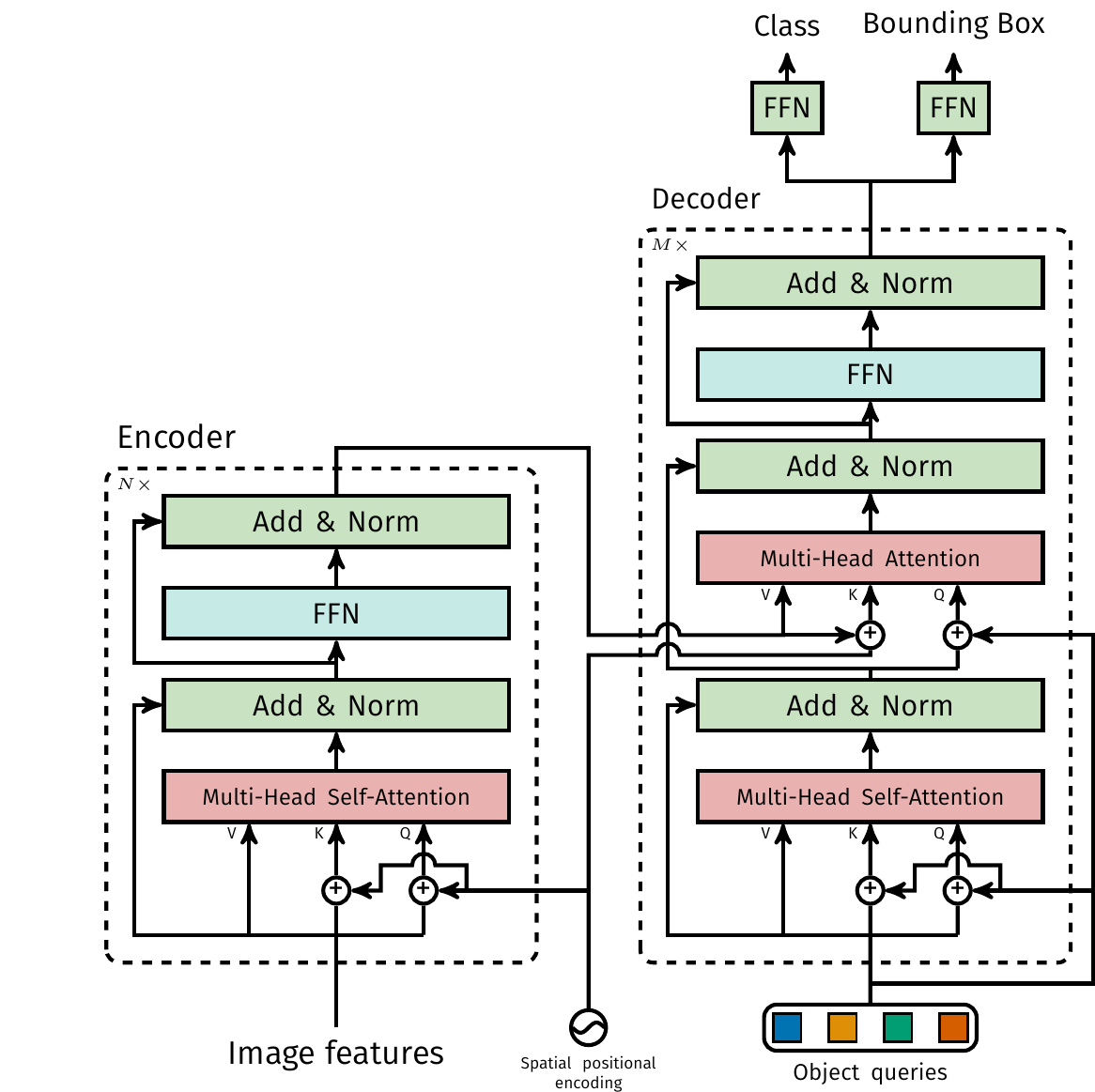}
    \caption{Architecture of \detr's transformer. Please, see Section \ref{sec:detailed_arch} for details.
    }
    \label{fig:transformer}
\end{figure}

The detailed description of the transformer used in \detr, with positional encodings passed at every attention layer, is given in Fig.~\ref{fig:transformer}.
Image features from the CNN backbone are passed
through the transformer encoder, together with spatial positional encoding
that are added to queries and keys at every multi-head self-attention layer.
Then, the decoder receives queries (initially set to zero),
output positional encoding (object queries), and encoder memory, and produces the final
set of predicted class labels and bounding boxes through multiple
multi-head self-attention and decoder-encoder attention.
The first self-attention layer in the first decoder layer can be skipped.
\oldnew{We use \detr-256 with 8 heads, and \detr-512 with 16 heads.}{}

\subsubsection{Computational complexity}
Every self-attention in the encoder has complexity $\mathcal{O}(\dmodel^2 HW+\dmodel(HW)^2)$: $\mathcal{O}(\dk \dmodel)$ is the cost of computing a single query/key/value embeddings (and $M\dk = \dmodel$), while $\mathcal{O}(\dk(HW)^2)$ is the cost of computing the attention weights for one head. Other computations are negligible. In the decoder, each self-attention is in $\mathcal{O}(\dmodel^2 N+\dmodel N^2)$, and cross-attention between encoder and decoder is in $\mathcal{O}(\dmodel^2(N+HW) + \dmodel NHW)$, which is much lower than the encoder since $N \ll HW$ in practice.
\subsubsection{FLOPS computation}
Given that the FLOPS for Faster R-CNN depends on the number of proposals in the image, we report the average number of FLOPS for the first 100 images in the COCO 2017 validation set.
We compute the FLOPS with the tool \texttt{flop\_count\_operators} from Detectron2 \cite{wu2019detectron2}. We use it without modifications for Detectron2 models, and extend it to take batch matrix multiply (\texttt{bmm}) into account for \detr models.

\subsection{Training hyperparameters}
\label{sec:hyperparameters}
We train \detr using AdamW~\cite{Loshchilov2017DecoupledWD} with improved weight
decay handling, set to $10^{-4}$.
We also apply gradient clipping, with a maximal gradient norm of $0.1$.
The backbone and the transformers are treated slightly differently, we now discuss the details for both.
\subsubsection{Backbone}
ImageNet pretrained backbone ResNet-50 is imported from Torchvision, discarding the last classification layer.
Backbone batch normalization weights and statistics are frozen during training,
following widely adopted practice in object detection.
We fine-tune the backbone using learning rate of $10^{-5}$. We observe that having the backbone learning rate roughly an order of magnitude smaller than the rest of the network is important to stabilize training, especially in the first few epochs.
\subsubsection{Transformer}
We train the transformer with a learning rate of $10^{-4}$.
Additive dropout of $0.1$ is applied after every multi-head attention and FFN before layer normalization.
The weights are randomly initialized with Xavier initialization.
\subsubsection{Losses} We use linear combination of $\ell_1$ and GIoU losses for bounding box regression with $\lambda_{\rm L1} = 5$ and $\lambda_{\rm iou} = 2$ weights respectively. All models were trained with $N=100$ decoder query slots.
\subsubsection{Baseline} Our enhanced Faster-RCNN+ baselines use GIoU~\cite{Rezatofighi_2018_CVPR} loss along with the standard $\ell_1$ loss for bounding box regression. We performed a grid search to find the best weights for the losses and the final models use only GIoU loss with weights $20$ and $1$ for box and proposal regression tasks respectively. For the baselines we adopt the same data augmentation as used in \detr and train it with 9$\times$ schedule (approximately 109 epochs). All other settings are identical to the same models in the Detectron2 model zoo~\cite{wu2019detectron2}.

\subsubsection{Spatial positional encoding}

Encoder activations are associated with corresponding spatial positions of image features. In our model we use a fixed absolute encoding to represent these spatial positions. We adopt a generalization of the original Transformer~\cite{Vaswani2017AttentionIA} encoding to the 2D case~\cite{Parmar2018ImageT}. Specifically, for both spatial coordinates of each embedding we independently use $\frac{d}{2}$ sine and cosine functions with different frequencies. We then concatenate them to get the final $d$ channel positional encoding.

\subsection{Additional results}

\begin{figure}[t!]
    \centering
     \begin{subfigure}[t]{\textwidth}
       \centering
       \includegraphics[width=.32\textwidth]{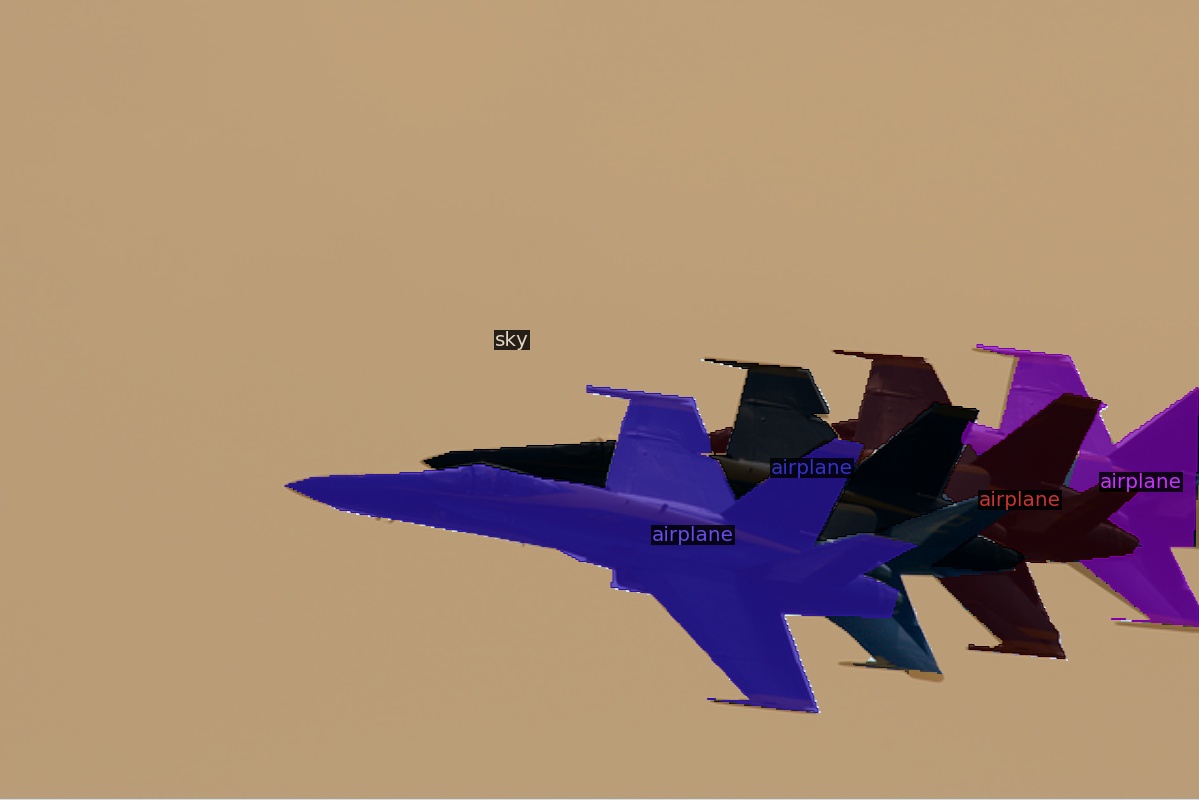}
       \includegraphics[width=.32\textwidth]{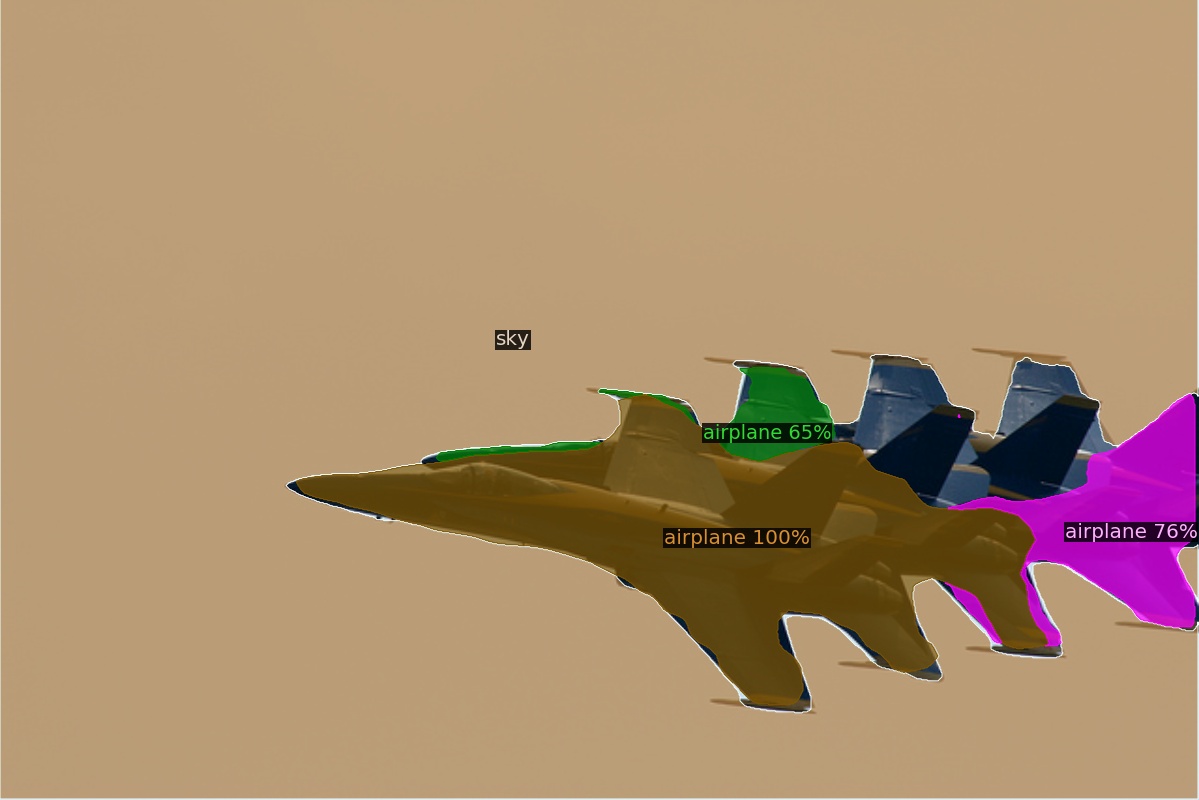}
       \includegraphics[width=.32\textwidth]{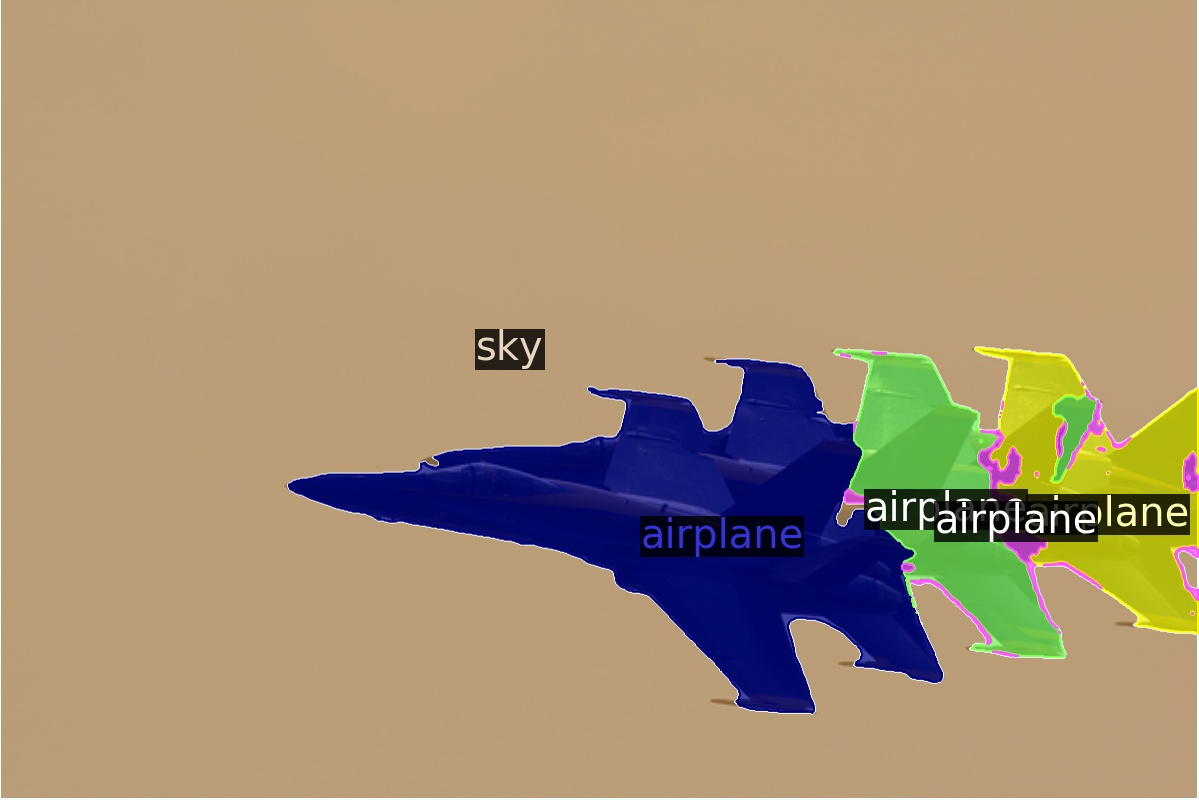}

       \caption{Failure case with overlapping objects. PanopticFPN misses one
         plane entirely, while \detr fails to accurately segment 3 of them.}
     \end{subfigure}
    \ \\
    \begin{subfigure}[t]{\textwidth}
        \centering
        \includegraphics[width=.32\textwidth]{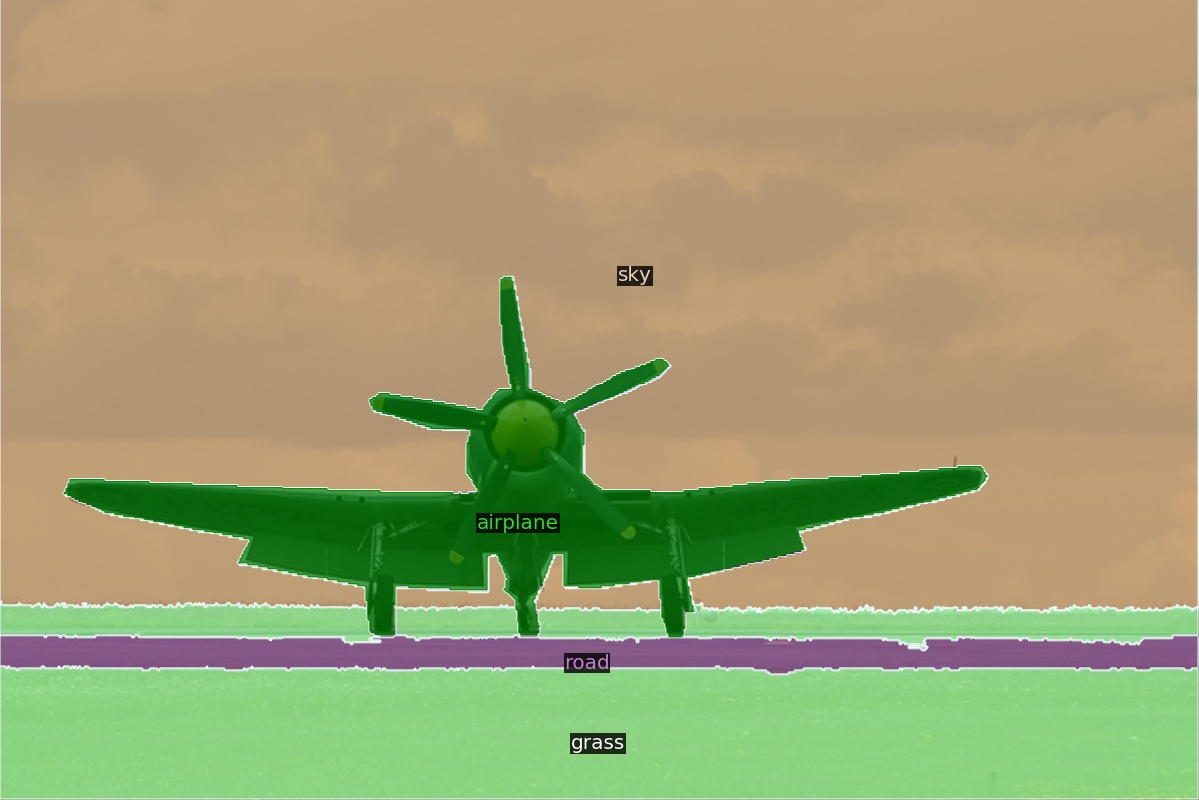}
        \includegraphics[width=.32\textwidth]{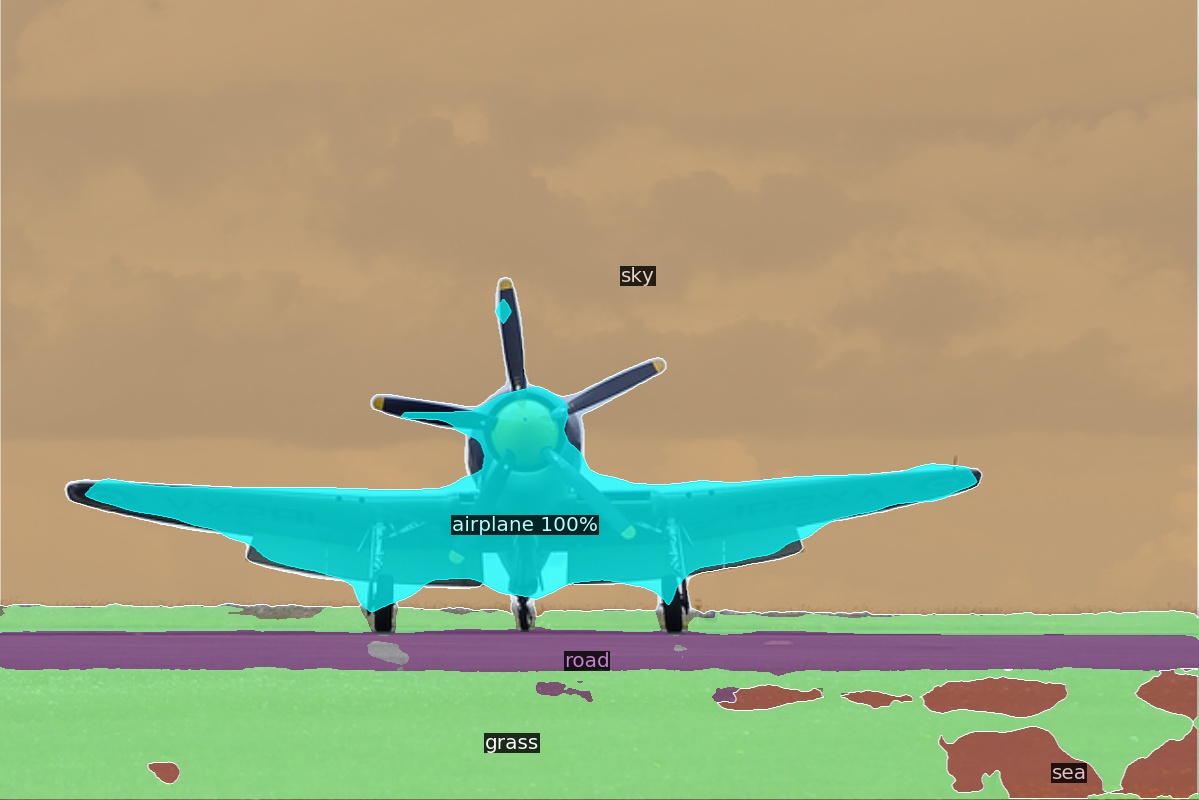}
        \includegraphics[width=.32\textwidth]{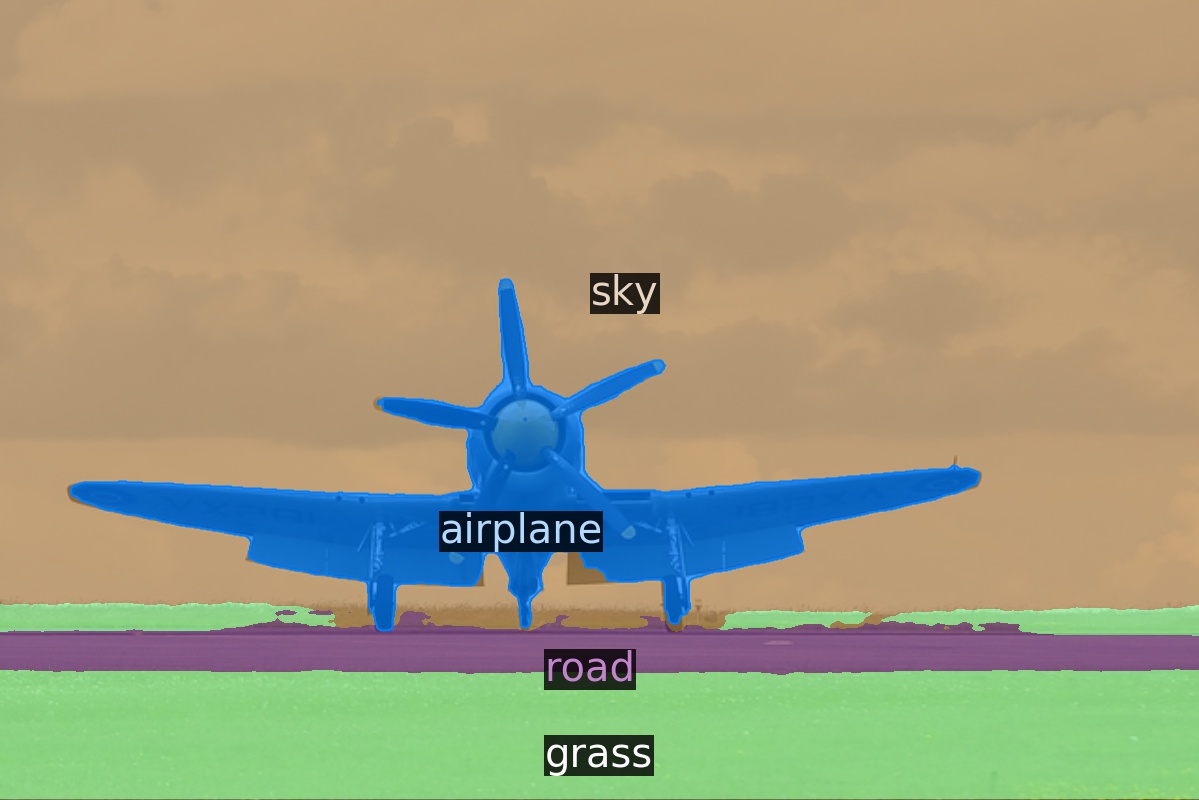}

        \caption{\texttt{Things} masks are predicted at full resolution, which allows sharper boundaries than PanopticFPN}
    \end{subfigure}%
\caption{Comparison of panoptic predictions. From left to right: Ground truth, PanopticFPN with ResNet 101, DETR with ResNet 101}
\label{fig:pano_quali_supl}
\end{figure}

Some extra qualitative results for the panoptic prediction of the \detr-R101 model are
shown in Fig.\ref{fig:pano_quali_supl}.

\subsubsection{Increasing the number of instances}
\begin{figure}
    \centering\small
    \includegraphics[ width=.6\columnwidth]{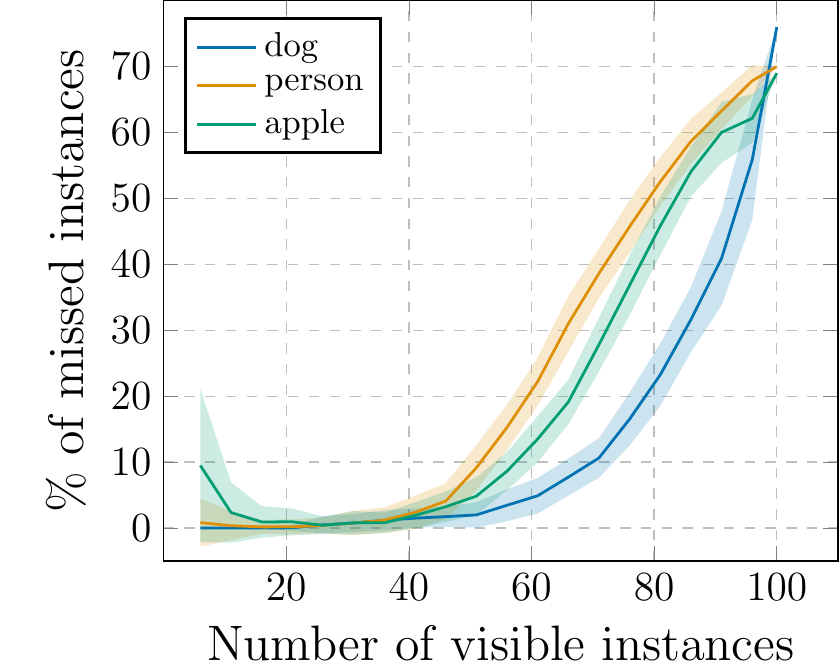}
    \caption{Analysis of the number of instances of various classes missed by \detr depending on how many are present in the image. We report the mean and the standard deviation. As the number of instances gets close to 100, \detr starts saturating and misses more and more objects}
    \label{fig:instances}
\end{figure}
By design, \detr cannot predict more objects than it has query slots, i.e. 100 in our experiments. In this section, we analyze the behavior of \detr when approaching this limit. We select a canonical square image of a given class, repeat it on a $10\times 10$ grid, and compute the percentage of instances that are missed by the model. To test the model with less than 100 instances, we randomly mask some of the cells.
This ensures that the absolute size of the objects is the same no matter how many are visible. To account for the randomness in the masking, we repeat the experiment 100 times with different masks. The results are shown in Fig.\ref{fig:instances}. The behavior is similar across classes, and while the model detects all instances when up to 50 are visible, it then starts saturating and misses more and more instances. Notably, when the image contains all 100 instances, the model only detects 30 on average, which is less than if the image contains only 50 instances that are all detected. The counter-intuitive behavior of the model is likely because the images and the detections are far from the training distribution. 

Note that this test is a test of generalization out-of-distribution by design, since there are very few example images with a lot of instances of a single class. It is difficult to disentangle, from the experiment, two types of out-of-domain generalization: the image itself vs the number of object per class. But since few to no COCO images contain only a lot of objects of the same class, this type of experiment represents our best effort to understand whether query objects overfit the label and position distribution of the dataset. Overall, the experiments suggests that the model does not overfit on these distributions since it yields near-perfect detections up to 50 objects.

\subsection{PyTorch inference code}

To demonstrate the simplicity of the approach, we include inference code
with PyTorch and Torchvision libraries in Listing~\ref{listing:code}.
The code runs with Python 3.6+, PyTorch 1.4 and Torchvision 0.5. Note that it does not support batching, hence it is suitable only for inference or training with DistributedDataParallel with one image per GPU. Also note that for clarity, this code uses learnt positional encodings in the encoder instead of fixed, and positional encodings are  added to the input only instead of at each transformer layer. Making these changes requires going beyond PyTorch implementation of transformers, which hampers readability. The entire code to reproduce the experiments will be made available before the conference.

\begin{listing}[ht]
\begin{minted}{python}
import torch
from torch import nn
from torchvision.models import resnet50

class DETR(nn.Module):

    def __init__(self, num_classes, hidden_dim, nheads,
                 num_encoder_layers, num_decoder_layers):
        super().__init__()
        # We take only convolutional layers from ResNet-50 model
        self.backbone = nn.Sequential(*list(resnet50(pretrained=True).children())[:-2])
        self.conv = nn.Conv2d(2048, hidden_dim, 1)
        self.transformer = nn.Transformer(hidden_dim, nheads,
                                          num_encoder_layers, num_decoder_layers)
        self.linear_class = nn.Linear(hidden_dim, num_classes + 1)
        self.linear_bbox = nn.Linear(hidden_dim, 4)
        self.query_pos = nn.Parameter(torch.rand(100, hidden_dim))
        self.row_embed = nn.Parameter(torch.rand(50, hidden_dim // 2))
        self.col_embed = nn.Parameter(torch.rand(50, hidden_dim // 2))

    def forward(self, inputs):
        x = self.backbone(inputs)
        h = self.conv(x)
        H, W = h.shape[-2:]
        pos = torch.cat([
            self.col_embed[:W].unsqueeze(0).repeat(H, 1, 1),
            self.row_embed[:H].unsqueeze(1).repeat(1, W, 1),
        ], dim=-1).flatten(0, 1).unsqueeze(1)
        h = self.transformer(pos + h.flatten(2).permute(2, 0, 1),
                             self.query_pos.unsqueeze(1))
        return self.linear_class(h), self.linear_bbox(h).sigmoid()

detr = DETR(num_classes=91, hidden_dim=256, nheads=8, num_encoder_layers=6, num_decoder_layers=6)
detr.eval()
inputs = torch.randn(1, 3, 800, 1200)
logits, bboxes = detr(inputs)

\end{minted}
\caption{DETR PyTorch inference code. For clarity it uses learnt positional encodings in the encoder instead of fixed, and positional encodings are  added to the input only instead of at each transformer layer. Making these changes requires going beyond PyTorch implementation of transformers, which hampers readability. The entire code to reproduce the experiments will be made available before the conference.}
\label{listing:code}
\end{listing}

%% file: preliminaries.tex
\subsection{Preliminaries: Multi-head attention layers}
\label{sec:mha} 

Since our model is based on the Transformer architecture, we remind here the general form of attention mechanisms we use for exhaustivity. The attention mechanism follows \cite{Vaswani2017AttentionIA}, except for the details of positional encodings (see Equation \ref{eq:posenc}) that follows \cite{Cordonnier2019OnTR}.

\subsubsection{Multi-head} The general form of \emph{multi-head attention} with $M$ heads of dimension $\dmodel$ is a function with the following signature (using  $\dk = \frac{\dmodel}{M}$, and giving matrix/tensors sizes in underbrace)
\begin{equation}
    \mhattn:\underbrace{\xq}_{\dmodel\times \Nq}, \underbrace{\xkv}_{\dmodel\times \Nkv}, \underbrace{\weit}_{M\times 3\times \dk \times \dmodel}, \underbrace{\proj}_{\dmodel\times \dmodel} \mapsto \underbrace{\xqout}_{\dmodel\times \Nq}
\end{equation}
where $\xq$ is the \emph{query sequence} of length $\Nq$, $\xkv$ is the \emph{key-value sequence} of length $\Nkv$ (with the same number of channels $\dmodel$ for simplicity of exposition), $\weit$ is the weight tensor to compute the so-called query, key and value embeddings, and $\proj$ is a projection matrix. The output is the same size as the query sequence. To fix the vocabulary before giving details, multi-head \emph{self-}attention ($\mhsattn$) is the special case $\xq = \xkv$, i.e. 
\begin{equation}
    \mhsattn(X, \weit, \proj) = \mhattn(X, X, \weit,\proj)\,.\label{eq:selfatt}
\end{equation}

The multi-head attention is simply the concatenation of $M$ single attention heads followed by a projection with $\proj$. The common practice \cite{Vaswani2017AttentionIA} is to use residual connections, dropout and layer normalization. In other words, denoting $\xqout = \mhattn(\xq, \xkv, \weit, \proj)$ and $\bar{\bar{X}}^{(q)}$ the concatenation of attention heads, we have
\begin{align}
    \xqbb&= [\attn(\xq, \xkv, \weit_1) ; ... ; \attn(\xq, \xkv, \weit_M)]\\
    \xqout &= {\rm layernorm}\big(\xq + {\rm dropout}(\proj\xqbb)\big)\,,\label{eq:layernormresdropout}
\end{align}
where [;] denotes concatenation on the channel axis. 

\subsubsection{Single head} An attention head with weight tensor $\weit'\in\Re^{3\times \dk\times \dmodel}$, denoted by $\attn(\xq, \xkv, \weit')$, depends on additional positional encoding $\posq \in\Re^{\dmodel\times \Nq}$ and $\poskv\in\Re^{\dmodel\times \Nkv}$. 
It starts by computing so-called query, key and value embeddings after adding the query and key positional encodings \cite{Cordonnier2019OnTR}:
\begin{equation}
    [\ques; \keys; \vals] = [\weit'_1 (\xq+\posq);\weit'_2 (\xkv+\poskv); \weit'_3 \xkv]
\end{equation}
where $T'$ is the concatenation of $T'_1,T'_2,T'_3$. The \emph{attention weights} $\alpha$ are then computed based on the softmax of dot products between queries and keys, so that each element of the query sequence attends to all elements of the key-value sequence ($i$ is a query index and $j$ a key-value index):
\begin{equation}
    \alpha_{i,j} = \frac{e^{\frac{1}{\sqrt{\dk}}\ques_i^T\keys_j}}{Z_i} \text{~~where~} Z_i = \sum_{j=1}^{\Nkv}  e^{\frac{1}{\sqrt{\dk}}\ques_i^T\keys_{j}}\,.\label{eq:posenc}
\end{equation}
In our case, the positional encodings may be learnt or fixed, but are shared across all attention layers for a given query/key-value sequence, so we do not explicitly write them as parameters of the attention. We give more details on their exact value when describing the encoder and the decoder. The final output is the aggregation of values weighted by attention weights:  The $i$-th row is given by 
$%
    \attn_i(\xq, \xkv, \weit') = \sum_{j=1}^{\Nkv} \alpha_{i,j} \vals_j
$.

\subsubsection{Feed-forward network (FFN) layers} The original transformer alternates multi-head attention and so-called FFN layers \cite{Vaswani2017AttentionIA}, which are effectively multilayer 1x1 convolutions, which have $Md$ input and output channels in our case. The FFN we consider is composed of two-layers of 1x1 convolutions with ReLU activations. There is also a residual connection/dropout/layernorm after the two layers, similarly to \eqref{eq:layernormresdropout}.

\subsection{Losses}

For completeness, we present in detail the losses used in our approach. All  losses are normalized by the number of objects inside the batch. Extra care must be taken for distributed training: since each GPU receives a sub-batch, %
it is not sufficient to normalize by the number of objects in the local batch, 
since in general the sub-batches are not balanced across GPUs. 
Instead, it is important to normalize by the total number of objects in all sub-batches.

\subsubsection{Box loss}
 Similarly to \cite{RomeraParedes2015RecurrentIS,ren2017end}, we use a soft version of Intersection over Union in our loss,  together with a $\ell_1$ loss on $\hb$:
 \begin{equation}
     \bloss{b_{\sigma(i)}, \hb_i} = \lambda_{\rm iou}\iouloss{b_{\sigma(i)}, \hb_i} + \lambda_{\rm L1}||b_{\sigma(i)}- \hb_i||_1\,,
     \label{eq:bloss}
 \end{equation}
 where $\lambda_{\rm iou}, \lambda_{\rm L1}\in\Re$ are hyperparameters and $\iouloss{\cdot}$ is the generalized IoU \cite{Rezatofighi_2018_CVPR}:
 \begin{equation}
     \iouloss{b_{\sigma(i)}, \hb_i} = 1 - \bigg(\frac{ | b_{\sigma(i)}\cap \hb_i|}{|b_{\sigma(i)}\cup \hb_i|} - \frac{|B(b_{\sigma(i)},\hb_i)  \setminus b_{\sigma(i)} \cup \hb_i|}{|B(b_{\sigma(i)},\hb_i)|}\bigg)\,.
 \end{equation}
 $|.|$ means ``area'', and the union and intersection of box coordinates are used as shorthands for the boxes themselves. The areas of unions or intersections are computed by $\min/\max$ of the linear functions of $b_{\sigma(i)}$ and $\hb_i$, which makes the loss sufficiently well-behaved for stochastic gradients. $B(b_{\sigma(i)},\hb_i)$ means the largest box containing $b_{\sigma(i)},\hb_i$ (the areas involving $B$ are also computed based on $\min/\max$ of linear functions of the box coordinates).
 
\subsubsection{DICE/F-1 loss~\cite{milletari2016v}}
The DICE coefficient is closely related to the Intersection over Union. If we denote by $\hat{m}$ the raw mask logits prediction of the model, and $m$ the binary target mask, the loss is defined as:
\begin{equation}
\diceloss{m, \hat{m}} = 1 - \frac{2 m\sigma(\hat{m}) + 1}{\sigma(\hat{m}) + m + 1}
\end{equation}
where $\sigma$ is the sigmoid function. This loss is normalized by the number of objects.